\documentclass[10pt,twocolumn,letterpaper]{article}

\usepackage[pagenumbers]{iccv} 
\usepackage{enumitem}

\newcommand{\red}[1]{{\color{red}#1}}
\newcommand{\todo}[1]{{\color{red}todo}}

\usepackage{url}
\usepackage[utf8]{inputenc} 
\usepackage[T1]{fontenc}    
\usepackage{booktabs}       
\usepackage{amsfonts}       
\usepackage{nicefrac}       
\usepackage{microtype}      
\newcommand{\cmark}{\ding{51}}%
\newcommand{\xmark}{\ding{55}}%
\usepackage{makecell}
\usepackage{setspace}
\usepackage{multirow}
\usepackage{graphicx}
\usepackage{subcaption}
\usepackage{pifont}%
\usepackage{xspace}
\usepackage{float}
\usepackage{mdframed}

\floatstyle{plaintop}
\restylefloat{table}
\usepackage[export]{adjustbox}
\newcommand{\dataset}{\textsc{Minerva}}
\newcommand{\metric}{MiRA}
\DeclareCaptionLabelFormat{andtable}{#1~#2  \&  \tablename~\thetable}

\definecolor{iccvblue}{rgb}{0.21,0.49,0.74}
\usepackage[pagebackref,breaklinks,colorlinks,allcolors=iccvblue]{hyperref}

\def\paperID{} 
\def\confName{ICCV}
\def\confYear{2025}

\newcommand{\new}[1]{#1}

\title{\dataset{}: Evaluating Complex Video Reasoning}

\author{Arsha Nagrani$^{1}$, 
Sachit Menon$^{2}$\thanks{Work done during an internship at Google DeepMind}, 
Ahmet Iscen$^{1}$, 
Shyamal Buch$^{1}$, 
Ramin Mehran$^{1}$, Nilpa Jha$^{1}$, \\
Anja Hauth$^{1}$, Yukun Zhu$^{1}$, Carl Vondrick$^{2}$, Mikhail Sirotenko$^{1}$, Cordelia Schmid$^{1}$, Tobias Weyand$^{1}$
\\%
\small $^{1}$ Google DeepMind, $^{2}$ Columbia University}
\begin{document}

\newmdenv[backgroundcolor=gray!10, linecolor=black, linewidth=0.5pt, roundcorner=4pt, innerleftmargin=2pt, innerrightmargin=2pt, innertopmargin=5pt, innerbottommargin=5pt]{exampleprompt}

\maketitle
\begin{abstract} 
Multimodal LLMs are turning their focus to video benchmarks, however most video benchmarks only provide \textit{outcome} supervision, with no intermediate or interpretable reasoning steps. This makes it challenging to assess if models are truly able to combine perceptual and temporal information to reason about videos, or simply get the correct answer by chance or by exploiting linguistic biases. To remedy this, we provide a new video reasoning dataset called \dataset{} for modern multimodal models. Each question in the dataset comes with 5 answer choices, as well as detailed, hand-crafted reasoning traces. Our dataset is multimodal, diverse in terms of video domain and length, and consists of complex multi-step questions. Extensive benchmarking shows that our dataset provides a challenge for frontier open-source and proprietary models. We perform fine-grained error analysis to identify common failure modes across various models, and create a taxonomy of reasoning errors. We use this to explore both human and LLM-as-a-judge methods for scoring video reasoning traces, and find that failure modes are primarily related
to temporal localization, followed by visual perception errors, as opposed to logical or completeness errors. \new{The dataset, along with questions, answer candidates and reasoning traces will be publicly available under} 
\href{https://github.com/google-deepmind/neptune?tab=readme-ov-file#minerva}{https://github.com/google-deepmind/neptune?tab=readme-ov-file\#minerva}.
\end{abstract}    
\section{Introduction}
\label{sec:intro}
\begin{figure*}
\centering
  \includegraphics[width=0.95\linewidth]{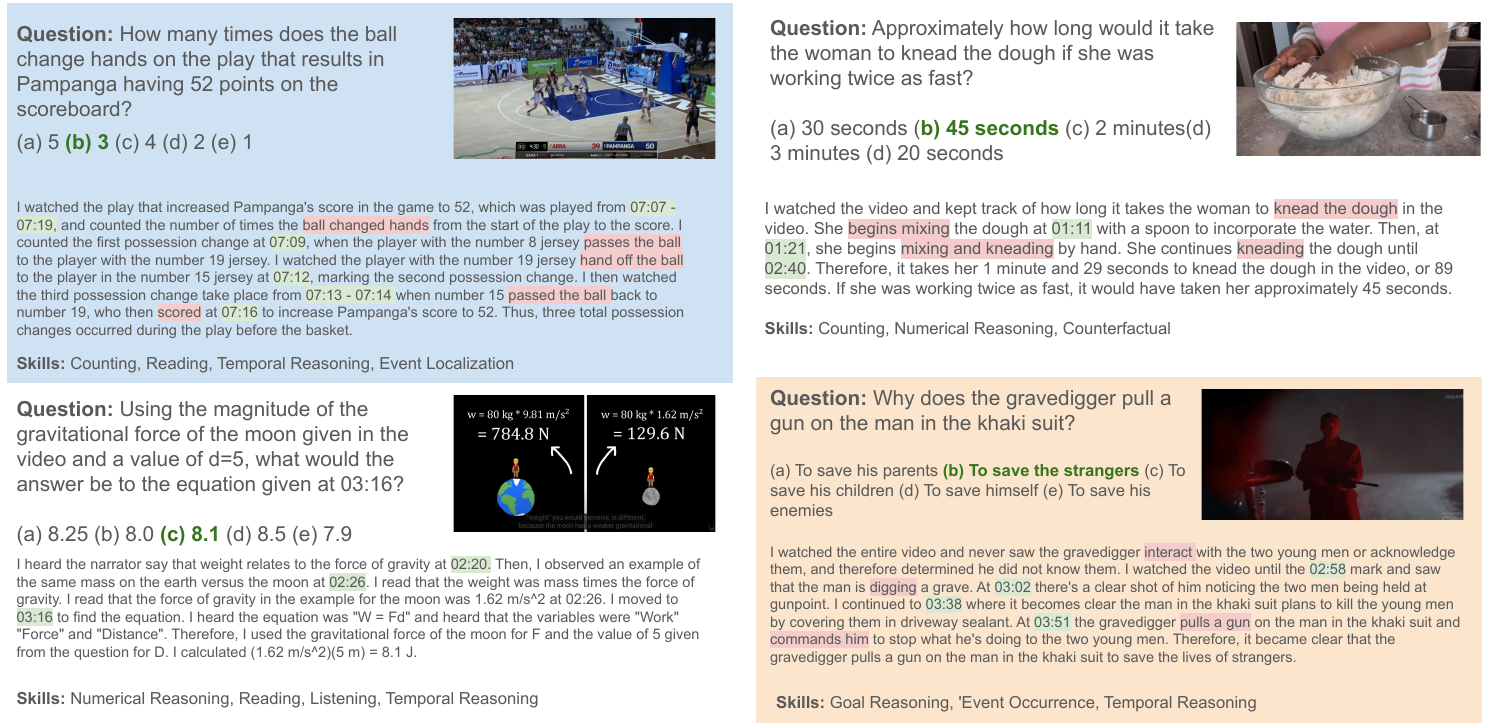}
   
  \caption{\textbf{Examples from \dataset{}:} We introduce \dataset{}, a complex video question-answering dataset. Unlike existing video datasets, the answer to each question is accompanied by a detailed \textit{reasoning trace}, which outlines the steps required to come to the answer. Videos cover multiple domains such as (clockwise) -- sports, cooking, short films and science lectures. Reasoning traces are detailed, including \textit{timestamps} (highlighted in green) and \textit{key actions} (highlighted in pink). We show a single frame from each video.}
  \vspace{-2mm}
  \label{fig:dataset-examples}
  \end{figure*}
 
Video understanding has long been a holy grail for the field of computer vision. Video benchmarks have evolved from testing action recognition on short clips~\cite{yu2019activitynet,kay2017kinetics,goyal2017something} 
to more recent benchmarks testing advanced capabilities over longer time horizons (LVBench~\cite{wang2024lvbench}, VideoMME~\cite{fu2024video}, Neptune~\cite{nagrani2024neptune}, and others). The development of the latter has been supercharged by the rise of multimodal large language models~\cite{achiam2023gpt, reid2024gemini, bai2025qwen2}, which perform complex video tasks on videos up to an hour long~\cite{wang2024lvbench}), and are steadily approaching human performance on these recent benchmarks~\cite{nagrani2024neptune,fu2024video}.

While these datasets aim to measure complex video understanding capabilities, they do not provide further insight into \textit{how} or \textit{why} models succeed or fail beyond the correctness of their final answers. One way to inspect model failures is by exposure to step-by-step rationales generated by the model.
Attempts to improve these rationales have led to a paradigm shift towards models (particularly in the text domain), that spend more time ``thinking"~\cite{openai-reasoning} i.e.\ producing intermediate thoughts towards a final answer, often via additional compute at inference time~\cite{jones2021scaling,snell2024scaling}. 
`Thinking' or reasoning in the video domain, however, looks very different to reasoning in the text or coding domain. In the text domain, models can directly quote sections or edited versions of the input as thoughts -- these intermediates are easier to assess by humans, or by another language model. This is not the case for the video domain. The high dimensionality and multi-modality of video means that solving a video task often requires temporal localization, perceptually recognizing key objects, events and actions (from multiple modalities such as speech or frames) and applying logical reasoning to tie these intermediates all together. Complex queries in video question answering (videoQA) hence naturally require multi-step processes to solve, where each step may require a different skill and/or access to a different modality input. We henceforth refer to this multi-step process as a `reasoning trace' for videoQA. This includes the set of the steps required to solve the question, including perception and localization.

Yet, despite the fundamentally multi-step nature of this capability, existing video benchmarks only evaluate final answers: they only check the \textit{outcome} and not the \textit{reasoning}. It is not clear, however, if a model arrives at a correct answer due to a successful execution of key steps, pure chance, linguistic bias, or the process of elimination of answer choices. Conversely, if it fails to produce the correct answer, was it completely unable to solve the problem, or did it get close but ultimately make a mistake that changed the final answer? Using final answers alone makes it difficult to design datasets that will remain challenging but still provide signal; all the more pertinent as new models improve at breakneck speed.

In light of this, we release \dataset{}\footnote{Multimodal INterpretablE Reasoning Video Annotations. Minerva is also the Roman goddess of thinking and wisdom.} -- a benchmark for video reasoning which consists of question, answers and reasoning traces.
\dataset{} is challenging and complex -- every question requires multiple steps to solve, and even the best-performing frontier model (\new{Gemini 2.5 Pro Thinking}) achieves only \new{66.2\%} accuracy, \new{while humans are able to achieve 92.5\%}. The dataset is hence well positioned for hill-climbing and future assessment of multimodal models. Every question in \dataset{} requires complex reasoning using two or more skills (for example numerical reasoning, temporal reasoning, spatial navigation). Videos also span multiple domains (short films, sports, instructional videos etc), with various video lengths (from 2 minutes to \emph{over 1.5 hours}), making the dataset diverse.
For each question we also provide the hand-crafted, detailed reasoning trace, with the steps that are required to come to the correct answer. Unlike datasets that provide auxiliary information in a single format such as timestamps (LITA~\cite{huang2024lita}, CG-Bench~\cite{chen2024cgbench} + others), the reasoning trace is an unconstrained block of text - allowing flexibility. Each reasoning trace does include timestamps where necessary to refer to relevant sections of the video, but also describes key actions, objects, as well as outlines logical reasoning steps (see Fig. \ref{fig:dataset-examples} for examples). 

We benchmark both open-source and proprietary models on our dataset, and manually compare some of their reasoning traces to the ground truth reasoning annotations to identify failure modes. We use these failure modes to build a taxonomy of errors (in other words, a rubric) for video reasoning. This rubric is specific to the video domain, and highlights the following broad categories of errors -  (1) Perceptual Correctness, (2) Temporal Localization, (3) Logical Reasoning and (4) Completeness. We use this rubric to assess how both humans and LLMs grade reasoning outputs along these criteria when given the ground truth reasoning, illuminating where automatic graders correlate to humans and where they may not, highlighting directions for potential improvement for reference-based metrics. In addition, we also find in some cases that LLM performance on video reasoning improves simply by being prompted with the rubric, in rough analogy to `Let's think step by step'~\cite{kojima2022large} in text reasoning.

To summarize, we make the following contributions: (i) We introduce \dataset{}, a challenging video reasoning benchmark for LMMs consisting of 1,515 hand-crafted questions. For each question, we provide 5 answer choices, as well as detailed, manually-annotated reasoning traces; (ii) We evaluate multiple frontier models on \dataset{}, both open-source and proprietary, and perform ablations to show low textual bias and the importance of visual information; (iii) we perform fine-grained error analysis to identify common failure modes across various models, and create a taxonomy of reasoning errors. We find that failure modes are primarily related
to temporal localization and visual perception. Preliminary analysis of LLM-based metrics show areas for future work.

\section{Related Work}
\subsection{Recent VideoQA Datasets}
A number of valuable video datasets have been proposed recently. TemporalBench~\cite{Cai2024temporalbench} and PerceptionTest~\cite{patraucean2024perception} include a variety of tasks, such as video QA, captioning, and grounding, but use relatively short videos (most videos a couple of minutes). 
Similarly, CinePile~\cite{rawal2024cinepile} and VideoVista~\cite{li2024videovista} focus on short-form content (average length of 160 seconds). 
VideoVista~\cite{li2024videovista} is notable for its broad coverage of 19 understanding and 8 reasoning tasks.
These benchmarks, along with InfiniBench~\cite{Ataallah2024infinibench} (maximum 52-minute videos), and Neptune~\cite{nagrani2024neptune}, rely on semi-automatic pipelines using LLMs for annotation. In contrast, our dataset is entirely manually annotated.  
TVBench~\cite{Cores2024tvbench} specifically addresses the challenge of temporally difficult answer candidates in VideoQA, emphasizing the importance of visual information. 
{E}.{T}. bench~\cite{Liu2024etbench} focuses on multi-event and time-sensitive understanding, using human-annotated videos with a maximum length of 795 seconds. 
Video-MME~\cite{fu2024video} bridges the gap between short and long-form content, and datasets such as 
LVBench~\cite{wang2024lvbench} (average length 68 minutes) and 
HourVideo~\cite{Chandrasegaran2024hourvideo} ( egocentric videos, 20 to 120 minutes) extending to even longer videos. 
In contrast to these existing benchmarks, our work provides not only the final outputs but also human-annotated reasoning traces, enabling future evaluations to assess the model's reasoning process in addition to its accuracy.

\begin{figure*}[t]
\begin{minipage}{0.65\linewidth}

\centering
  \includegraphics[width=\linewidth]{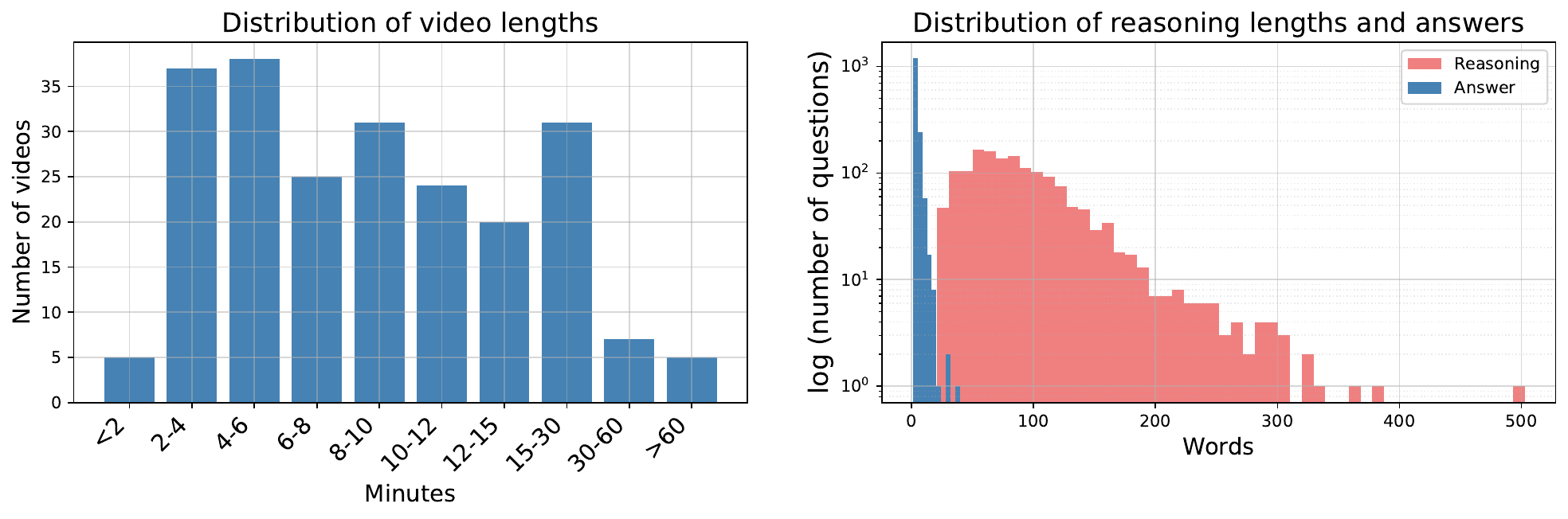}
\end{minipage}
\begin{minipage}{0.33\linewidth}
\centering
  \includegraphics[width=\linewidth]{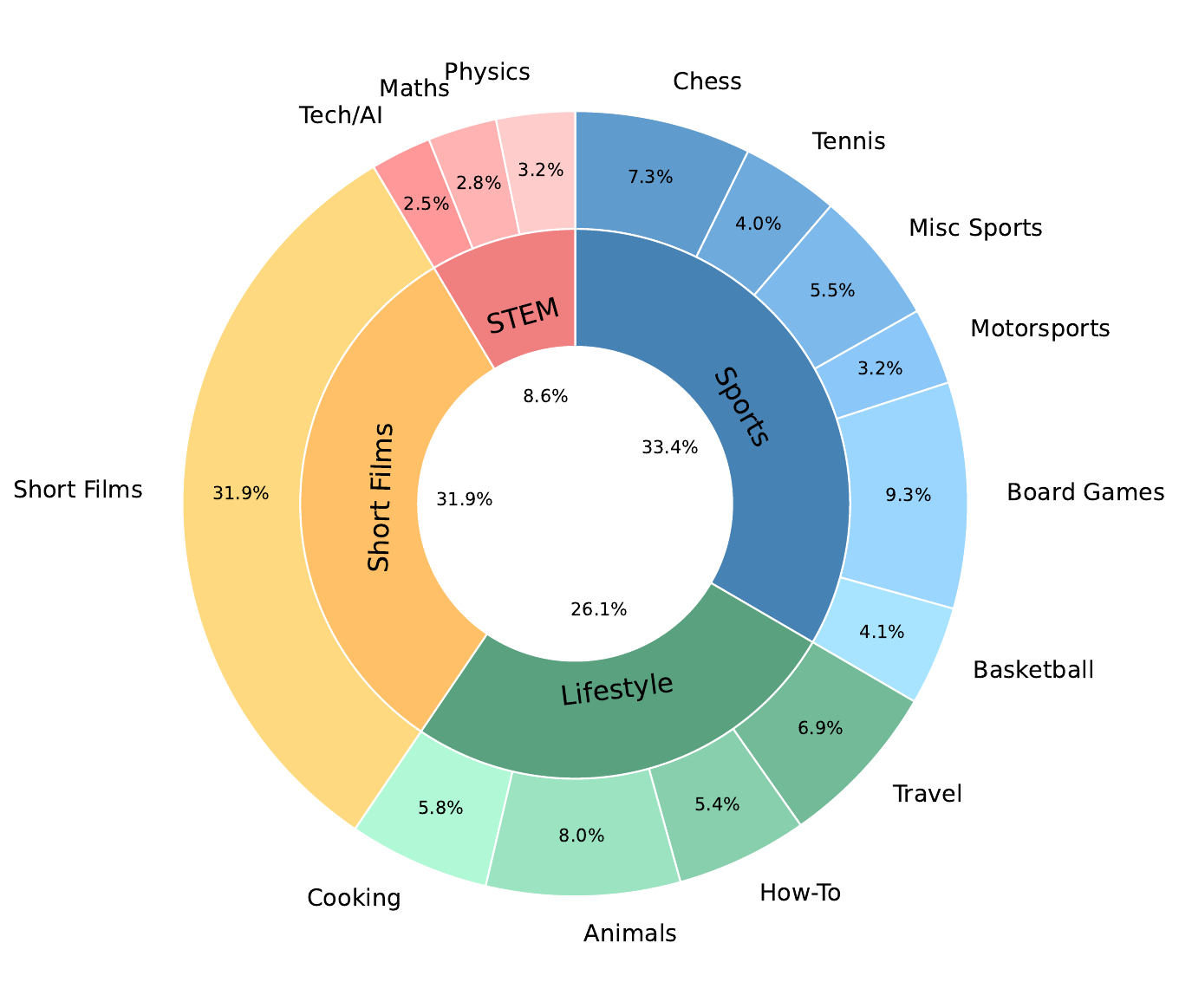}
\end{minipage}
\centering
\caption{\textbf{Dataset statistics.} Video lengths (left), lengths of answers and reasoning (middle), and domains (right). Videos cover a wide range of lengths, with some longer than 100 minutes. Every question comes with a reasoning trace which is long and detailed, mean number of words is 92 (middle). Domains are hand-selected to include videos that lend themselves well to complex reasoning questions. 
}
\label{fig:dataset-stats}
\end{figure*}
\subsection{Reasoning Traces for Video}
As discussed in the previous section, few datasets provide auxiliary dataset beyond final answers. ScienceQA~\cite{saikh2022scienceqa} consists of science questions about images, and provides the accompanying lectures, which when used for training results in increased performance with fewer data points.
Some recent works have aimed to develop automatic labeling pipelines to instead generate noisy auxiliary information at scale.
Visual CoT~\cite{shao2024visual} provides bounding box pseudolabels generated by off-the-shelf models for image VQA, while VideoEspresso~\cite{han2024videoespresso} does something similar for video, constructing a pipeline connecting different frozen models together to label bounding boxes for sparse key frames. VideoCoT~\cite{wang2024videocot} is perhaps the closest to our work, aiming to generate text-based chain of thought for videos; however, we note key differences here. The primary goal of VideoCoT is to present a semiautomatic pipeline using LLMs and VLMs to scale up for training purposes, while we present a fully manually annotated, high quality dataset for evaluation purposes. The videos in VideoCoT are all taken from established Kinetics700~\cite{kay2017kinetics}, making them shorter and limited in domain. Finally, their automated rationales tend to contain substantial information about the video that does not relate to the particular query, rather than providing specific reasoning for the given question.

\subsection{Reasoning Analysis}
With the growth of reasoning models, substantial work has been conducted into analyzing reasoning in text-only settings and remains an open problem \cite{mondorf2024beyond}. 
Previous works demonstrate that correctness of final answers alone often does not indicate correct reasoning, commonly due to hallucination of reasoning steps or answers \cite{hao2024llm,tyen2023llms,golovneva2022roscoe,prasad2023receval}. These works fall into (i) reference-based, which use additional human annotation as reference, and (ii) reference-free (evaluation without such additional information). Unsurprisingly, multiple works find that reference-based evaluation proves consistently more reliable \cite{deutsch2022limitations}, and that language models are poor evaluators of reasoning in isolation \cite{tyen2023llms, kamoi2024can,huang2023large}. 
In this work, we primarily focus on contributing a high-quality dataset with reference annotations for video reasoning, which may spur the development of and provide a comparison for further research into both reference-based and, by comparison, reference-free metrics for video reasoning. 

Earlier work in reference-based evaluation (before LLMs) explored rule based methods such as $n$-grams for summarization with ROUGE scores \cite{lin2004rouge} or for machine translation with BLEU \cite{papineni2002bleu}. ROSCOE \cite{golovneva2022roscoe} observed the need for evaluation of reasoning for early LLMs, inspired by chain-of-thought prompting \cite{wei2022chain}, using various customized metrics such as word vector similarity.
As LLMs advanced, they began to be used for evaluation themselves, such as via LLM-as-a-Judge~\cite{zheng2023judging}. 
MLLM as a judge~\cite{chen2024mllm} and LLaVA-Critic~\cite{xiong2024llava} show these capabilities in multimodal settings, but this has thus far been largely explored for the image-text domain. Unlike these works, we explore reference-based LLM-as-a-judge strategies for analysis of video reasoning traces. In both text and multimodal contexts, well-designed, itemized evaluation criteria for rating different model responses give substantially more consistent, well-aligned LLM scores~\cite{zheng2023judging,liu2023visual} -- we follow along these lines of work by providing a rubric for scoring video reasoning, and provide a preliminary analysis of its effectiveness. 
\section{\dataset{}}
We introduce \dataset{}, a videoQA dataset consisting of 1,515 challenging questions for videos of varying length. Each question has 5 answer options and a detailed reasoning trace accompanying the final answer (see Fig.~\ref{fig:dataset-examples} for examples). The key features of our dataset are: (i) \textbf{multistep, complex} questions. Each question requires multiple steps and multiple skills to solve, (ii) \textbf{multimodality,} with both frames and ASR needed for some questions, (iii) \textbf{high quality} -- the entire dataset is hand crafted by experienced annotators, and most importantly (iv) \textbf{intermediate reasoning} -- detailed, hand-crafted reasoning traces. These reasoning traces allow us to perform an analysis of where models are making errors, beyond just their final answer outputs. We do this via  \textbf{reference-based} analysis of model outputs, which can be performed solely in the text domain (and is therefore much cheaper than reference-free assessments). We showcase the value of this annotation via both human and LLM-based judgments of model outputs (in Sec. \ref{sec:analysis}), based on a taxonomy of errors and detailed rubric system. \\
\noindent\textbf{Statistics:} Statistics are provided in Fig. \ref{fig:dataset-stats}. The dataset covers a wide range of video lengths, with the longest video being 100 minutes while the shortest video is less than 2 minutes (mean of 12 minutes). There are multiple questions per video, with the distribution following a min/mean/max of 1/6.8/10 questions per video (223 videos in total). Reasoning traces are long and detailed, with the mean number of words in a reasoning trace being 92 (Fig. \ref{fig:dataset-stats}). 99.6\% of reasoning traces contain timestamps, with the mean number of timestamps per reasoning trace being 4.

\subsection{Dataset Construction}
As multimodal models continue to improve, it becomes increasingly non-trivial to come up with challenging questions. Questions should be complex, requiring multiple steps to solve, and not solvable with only the speech or external knowledge (full criteria provided to raters can be found in Appendix~\ref{appendix:question-raters}). Our dataset construction pipeline consists of the following steps: \\
\noindent\textbf{1. Video Selection:} We begin by selecting video domains from YouTube that lend themselves well to questions fulfilling the desiderata above. \\ 
\noindent\textbf{2. Manual Annotation:} Raters propose questions, answers and reasoning traces. \\
\noindent\textbf{3. Quality Review:} Questions are reviewed by other raters. \\
\noindent\textbf{4. Adversarial Filtering:} We attempt to mitigate textual bias using consensus from multiple frontier text-only models.
 
\subsubsection{Video Selection}
Video selection is non-trivial, as many videos online contain simple storylines, few shots, or are dominated by talking heads. To avoid this, we identify a (non-exhaustive) set of domains that lend themselves well to reasoning. These are described below (and shown in Fig. \ref{fig:dataset-stats}). \\
\noindent\textbf{Short Films:} 
We search for videos tagged as ``short film", which are complex, multi-event videos that typically have a story line. Short films tend to contain relationship and event arcs over varying time horizons. Note that this set does not include mainstream movies, so there is less chance of training data contamination or external knowledge bias. \\
\noindent\textbf{Sports and Board Games:} 
This set includes sports such as tennis, basketball, motorsports, hockey, cricket, and board games such as chess, scrabble and risk. These videos lend themselves well to questions that require reasoning against the rule set of the sport or game, as well as recognition of fine-grained actions and identification of player or piece positions on screen. \\
\noindent\textbf{Educational:} 
Inspired by reasoning works in the maths and textual domain, we identify a number of STEM lectures online. However this domain set makes up only 8\% of the dataset (Fig. \ref{fig:dataset-stats}, right), as it is challenging to find lectures where the speech does not dominate (and visual reasoning is important).    \\
\noindent\textbf{Lifestyle:} 
Our final domain category consists of VLOG-style lifestyle videos, including cooking, general how-to videos, as well as travel vlogs and pet videos. These are typically multi-e vent videos where events may follow a logical causal flow (\eg steps in a recipe). Travel and vlogging style videos also add egocentric viewpoints, and lend themselves well to questions requiring spatial reasoning. 
\subsubsection{Annotation and Quality Review}
Once videos are identified, raters then propose complex questions, answers, decoys, reasoning traces and label question types. The raters are instructed to attempt to propose questions such that each question requires at least two of the following skills: Temporal Reasoning,  
Counting, Cause and Effect, Goal Reasoning, Situational Awareness, Event Occurrence, State Changes, Reading (OCR), Listening (identifying a detail in the audio track), Spatial Perception, Numerical Reasoning (all math operations other than counting), Object Recognition, Counterfactual Reasoning (`what if', but with an objective outcome).
Each annotation is verified by at least one other rater (details in Appendix \ref{appendix:pipelining}). The annotations are also periodically reviewed by the papers' authors, who provide detailed feedback to the raters to ensure annotations are being provided in-line with guidelines. 
\subsubsection{Filtering for Quality and Biases}

Finally, we take several steps to address quality and potential for biases in the final dataset. We begin by filtering examples where the groundtruth annotated reasoning traces are of low quality or too short (<20 words). While raters are explicitly instructed to avoid proposing questions that can be solved from the ASR alone, we find (very few) examples that are possible to guess from text alone. We address these potential text-biases (both in terms of QAD-only and ASR-only) with adversarial filtering \cite{hsieh2023sugarcrepe,le2020adversarial}. Our filtering process consists of taking the \textit{consensus} agreement (in order to avoid discarding difficult questions that models may have answered correctly by chance) across a diverse range of open- and closed- source text-only baselines (Deepseek, GPT-4o, Gemini-flash-002 and Qwen2.5-VL) with only QAD and/or ASR inputs, and best practices in prior work \cite{wang2024lvbench}.
We provide additional details in the Appendix.

\begin{table}[t]
\centering
\small
\setlength{\tabcolsep}{2pt}
\begin{tabular}{llcr}
\toprule
\textbf{Method}  & \# of Frames & ASR & MCQ-Acc. \% \\
\midrule
GPT-4o~\citep{achiam2023gpt} & 64 & \cmark & 39.74 \\
GPT-4o~\citep{achiam2023gpt} & 250 (max) &  \xmark  & 44.82 \\
GPT-4o~\citep{achiam2023gpt} & 250 (max) & \cmark & 45.54 \\
\midrule
Gemini 2.0 Flash~\citep{reid2024gemini} & 64 & \cmark & 47.85 \\
Gemini 2.0 Flash~\citep{reid2024gemini} & 256 & \xmark & 49.70 \\
Gemini 2.0 Flash~\citep{reid2024gemini} & 256 & \cmark & 53.47\\
\midrule
Qwen-2.5VL~\cite{bai2025qwen2} & 64 & \cmark & 33.00 \\
Qwen-2.5VL~\cite{bai2025qwen2} & 768 (max) & \xmark & 34.39 \\
Qwen-2.5VL~\cite{bai2025qwen2} &  768 (max) & \cmark & 35.05 \\
\bottomrule
\end{tabular}
\caption{\textbf{Frame and ASR ablations on \dataset{}}.}
\label{tab:frame_asr_ablations}
\end{table}

\begin{table}[t]
\centering
\small
\setlength{\tabcolsep}{2pt}
\begin{tabular}{lrr}
\toprule
\textbf{Method}  & w/o ASR & w/ ASR \\
\midrule
Random & 20.00 & 20.00 \\
\midrule
GPT-4o~\citep{achiam2023gpt} & $\dagger$19.60 & 29.17 \\
Gemini 2.0 Flash~\citep{reid2024gemini} & 23.04 & 28.25 \\
Qwen-2.5VL~\cite{bai2025qwen2} & $\dagger$18.88 & 23.17 \\
DeepSeek~\cite{guo2025deepseek-r1} & 21.45 & 22.51 \\
\bottomrule
\end{tabular}
\caption{\textbf{Blind Baselines on \dataset{}.} $\dagger$Random performance.}
\label{tab:blind_baselines}
\end{table}

\section{Benchmarking}
\subsection{Models}
We benchmark a number of open-source and proprietary models, described below. \\
\noindent\textbf{Blind Baselines:} We first evaluate models using a text-only prompt in two settings. (i) The model is given only the question, answer and decoys (QAD baseline).
(ii) The model is additionally given an ASR transcript of the video (QAD+ASR baseline). This helps identify questions that can be answered by prior or commonsense knowledge, or from ASR alone without requiring visual information.\\
\noindent\textbf{Video Models:} 
We experiment with 3 of the best-performing open source VideoQA models, Qwen2.5-VL~\cite{bai2025qwen2}, VideoLLaMA3~\cite{zhang2025videollama} and InternVideo2.5~\cite{wang2022internvideo}, which represent the current state-of-the-art in long context open multimodal models (we do not evaluate older versions superseded by these, \eg VideoLLaMA2 or Qwen2).
For proprietary models, we experiment with the Gemini model family~\citep{reid2024gemini} from Google DeepMind, GPT-4o~\citep{achiam2023gpt} \new{GPT-4.1~\citep{openai-gpt4.1}} and  o1~\cite{openai-reasoning} from OpenAI, as well as Claude 3.5 Sonnet v2~\cite{claude35sonnetv2} from Anthropic. \new{o1 and Gemini 2.5~\cite{gemini2.5} are reasoning or `thinking' models, capable of reasoning through their thoughts before responding.}
For all models, we attempt to fit as many frames as possible - given API query limitations. We experiment with varying number of frames, and compare performance with and without ASR provided with the frames.
\subsubsection{Prompts}
\label{sec:promptsection}
We conduct an ablation on the impact of prompting styles on \dataset{} with our best model. We try out 3 styles of prompting: 
(i) asking the model to answer the question directly; (ii) asking the model to  `reason step by step'; and (iii) additionally providing the model with the \dataset{} rubric for video reasoning (described in Sec. \ref{sec:taxonomy}). Prompts are provided in the Appendix~\ref{appendix:model-prompts}. We then use the best result of this study to prompt all the other baselines. We tried to standardize prompts as far as possible, while attempting to incorporate author-provided model-specific prompts to give models the best possible advantage.
\subsubsection{Implementation Details}
For all models, we follow best practices and guidance from model/API authors. In general, we sample video frames uniformly from the videos and construct prompts by interleaving them with timestamps. Since many questions ask about specific times in the speech and the video, we also interleave the ASR with timestamps at five second intervals. We prompt models to provide their answer choice in a machine-readable format for parsing; if the model fails to provide a response or it is not readily parse-able, then we assign a random answer choice for computing MCQ accuracy. We provide detailed settings in the appendix.

\begin{figure*}[htbp]
    \centering
    \begin{subfigure}{0.37\textwidth}
        \centering
        \includegraphics[width=\textwidth]{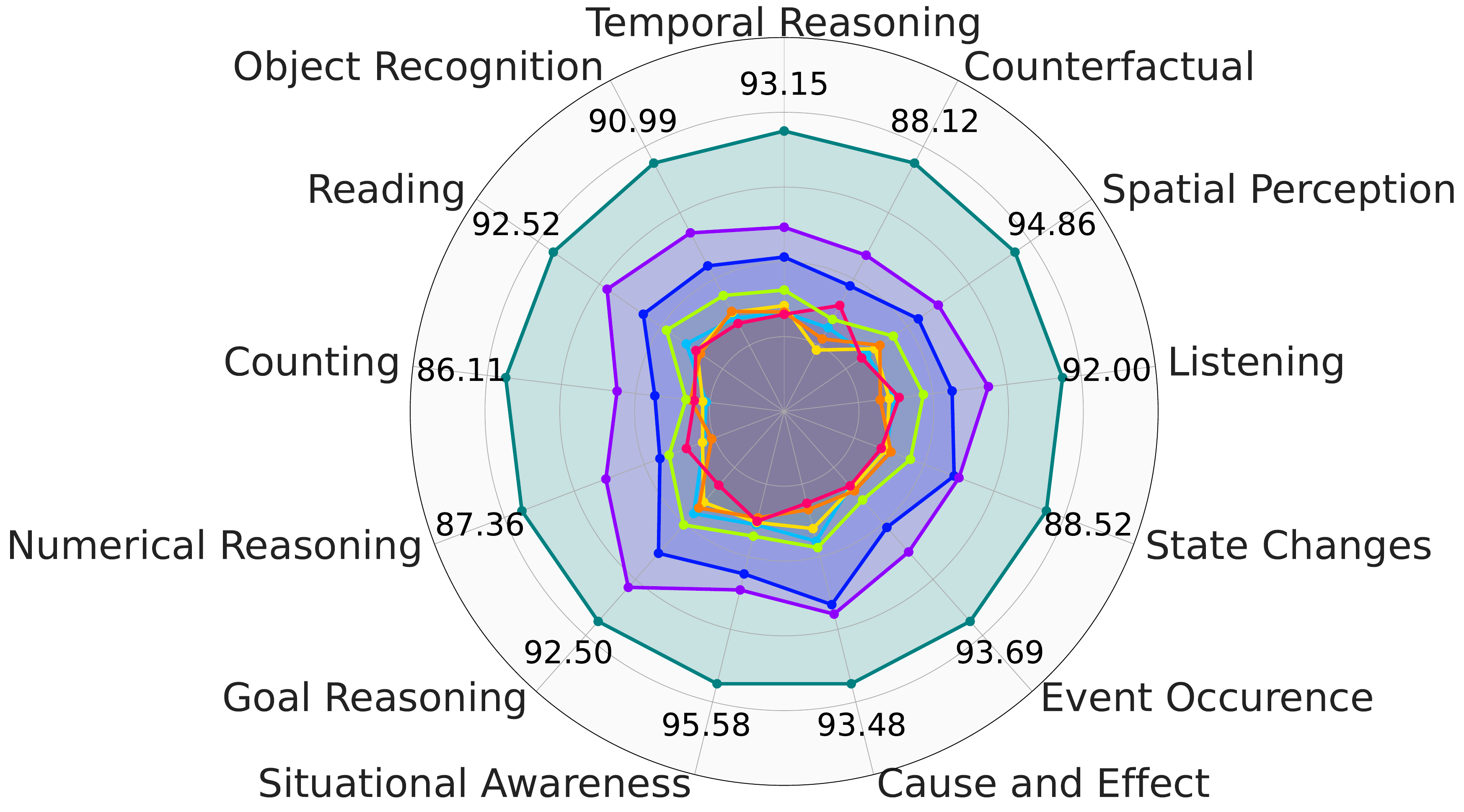}
        \caption{}
        \label{fig:figure3}
    \end{subfigure}
    \begin{subfigure}{0.29\textwidth}
        \centering
        \includegraphics[width=\textwidth]{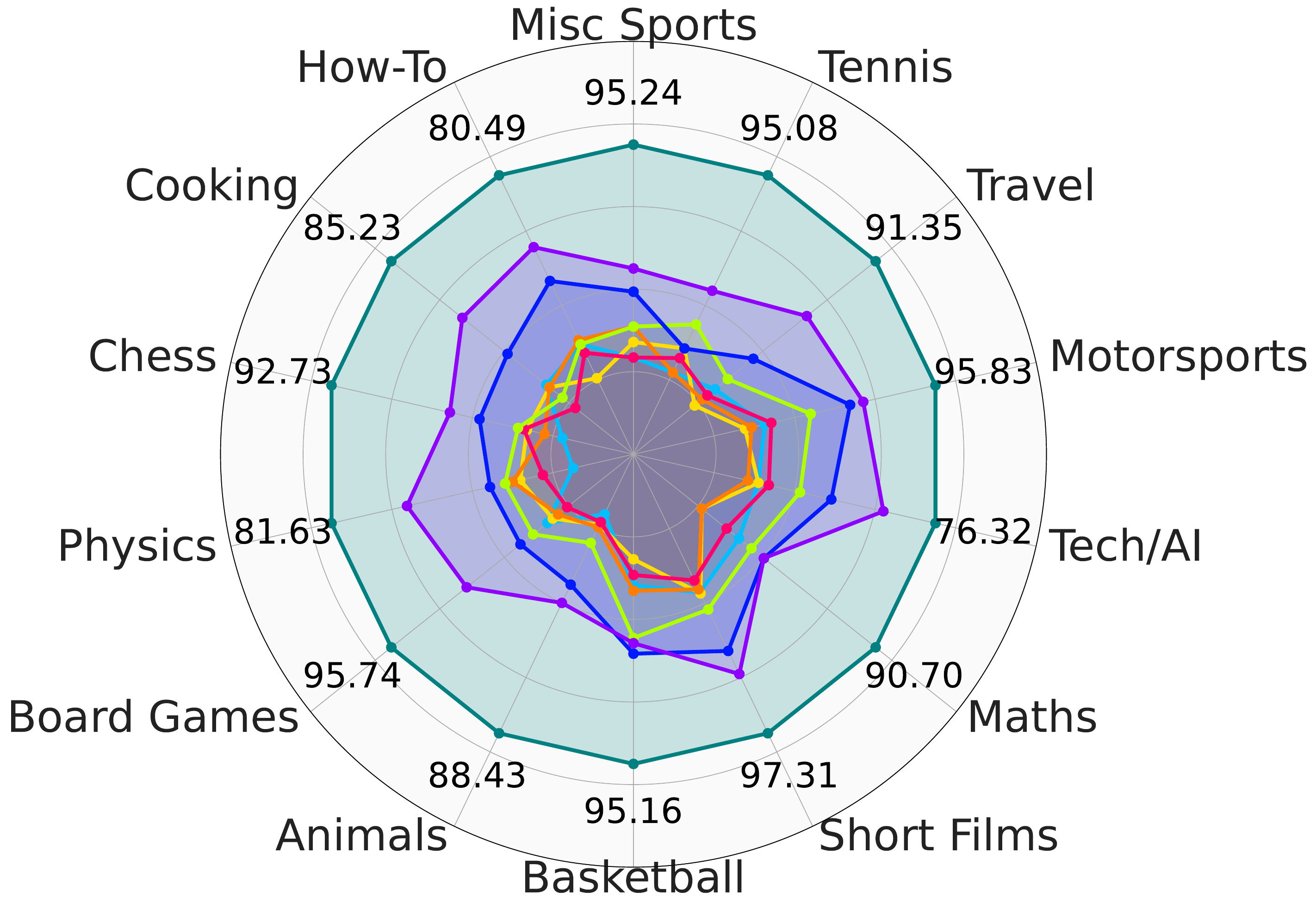}
        \caption{}
        \label{fig:figure1}
    \end{subfigure}%
    \begin{subfigure}{0.33\textwidth}
        \centering
        \includegraphics[width=\textwidth]{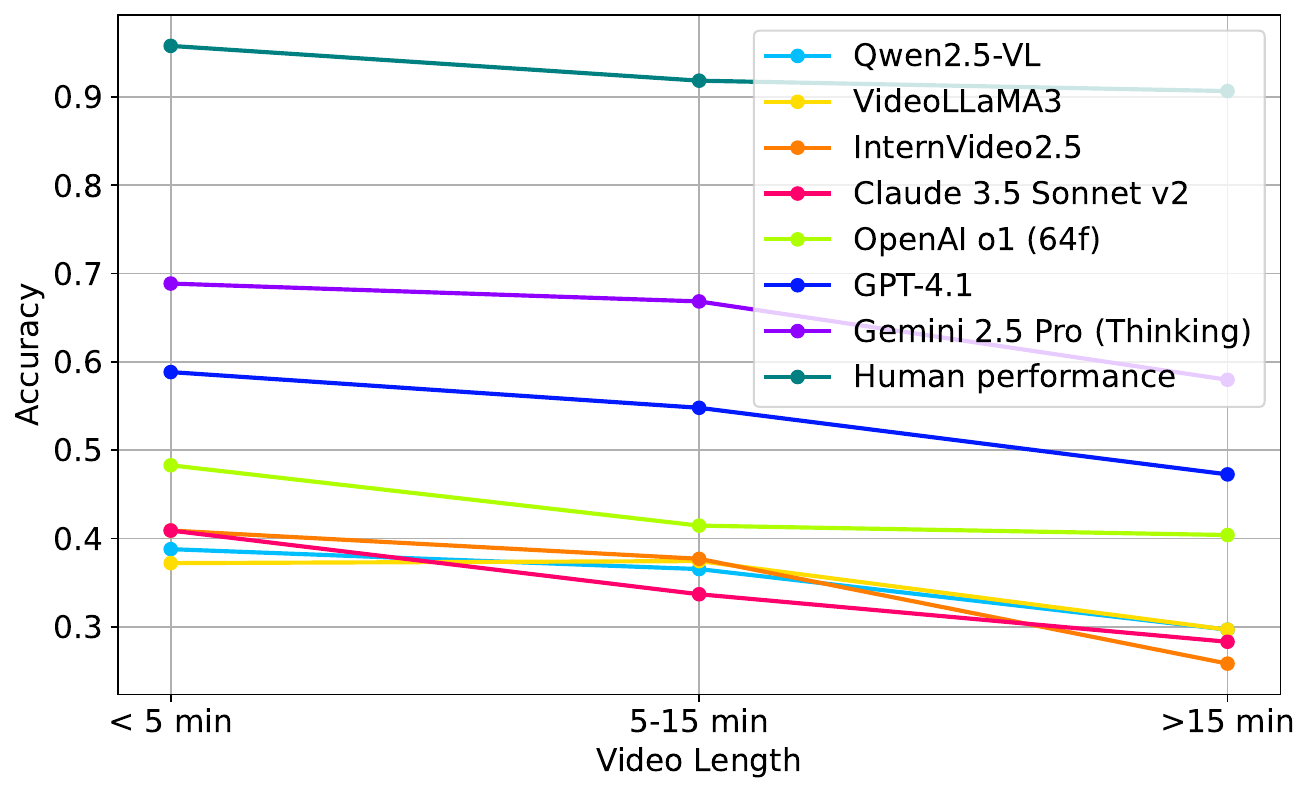}
        \caption{}
        \label{fig:figure2}
    \end{subfigure}%
    \caption{\textbf{Model \new{and human} accuracy (MCQ) broken down by (a) skill, (b) video domain, (c) video length.} Note that questions can belong to more than one skill for (a). We provide a common legend for all 3 plots (in the right). Best viewed in color and with zoom. Axes in the radar charts are scaled to the highest accuracy per dimension. The full results for (a) are provided in Tab.~\ref{tab:mcq_scores_by_skill} and for (b) in Tab.~\ref{tab:mcq_scores_by_category} in the appendix.}
    \label{fig:finegrained-acc}
\end{figure*}
  
\subsection{MCQ performance}

\noindent\textbf{Modality Ablations:} QAD-only and QAD+ASR-only baselines with no visual information are provided in Tab.~\ref{tab:blind_baselines}, with further modality ablations in Tab.~\ref{tab:frame_asr_ablations}. All QAD-only baselines get close to chance performance, indicating that the decoy answer choices do not offer cues to the correct answer.
ASR-only baselines using 4 powerful language models obtain extremely low performance compared to those with frames, showcasing the necessity of visual information for \dataset{}. We also note that for all 3 models in Table \ref{tab:frame_asr_ablations}, increasing the number of frames from 64 frames leads to an increase in performance as well, with ASR providing complementary gains. Finer-grained analysis on the impact of varying the number of frames provided for inference can be found in Fig.~\ref{fig:appendix-frame-ablations} in the appendix. \\
\noindent\textbf{Scoring final answers (MCQ):} We provide the MCQ results in Tab.~\ref{tab:results}. Contrary to prior work \cite{nagrani2024neptune}, we observe that the gap between open-source and proprietary models has narrowed, with Qwen2.5-VL and InternVideo2.5 outperforming Claude Sonnet. \new{Gemini 2.5 Pro Thinking} sets the state-of-the-art in the dataset at \new{66.2\%}. With peak performance \new{still far from human performance}, we hope \dataset{} will be a challenging benchmark to measure progress on video understanding.\\
\new{\noindent\textbf{Thinking Ablations:}} \new{We experiment with the thinking mode of the recently released Gemini 2.5 models~\cite{gemini2.5}. Results are provided in Tab.~\ref{tab:thinking-results}. We note that turning thinking mode on improves the performance of the model, with improvements more pronounced as the number of frames increases. This suggests that the ability to reason about long video content is necessary to achieve high accuracy scores on \textsc{Minerva}.} \\ 
\noindent\textbf{Results by Skill/Domain/Length:} Model performance by (a) skill, (b) video domain, and (c) video length is provided in Fig.~\ref{fig:finegrained-acc}. (We note that each question is tagged with multiple skills by construction, rather than each being associated with only one.) Models struggle most with counting questions, followed by counterfactuals and state changes. \new{Gemini 2.5 Pro performs highest across all skills, with GPT-4.1 performing similar to Gemini 2.5 Pro for state changes and cause and effect.} All models perform well in the domain of short films, while models struggle most on math videos. \new{Humans perform well across all skills, with the lowest performance for counting, and perform well across all domains except for Tech/AI.} The performance of all video models degrades as videos get longer, similar to reports in prior work~\cite{fu2024video}. \new{Human performance is also highest for videos less than 5 minutes long (95.8\%).} \\
\noindent\textbf{Prompt Ablations:} Results are provided in Table \ref{tab:prompt-ablation}. We find that asking the model to perform step-by-step reasoning rather than directly producing an answer results in a significant boost to MCQ accuracy. What is interesting however, is that explicitly providing the rubric in the prompt improves the final score even further (the reasoning outputs also improve, as shown by an automatic LLM judge (\metric{}) which is described in Sec. \ref{sec:taxonomy}). Note that this improvement comes with minimal extra inference-time compute (no multiple calls needed), and our rubric was designed to be as general as possible (does not contain any few-shot examples specific to the dataset). This suggests that asking models to provide reasoning along the four axes we identified in the rubric for video can actually improve final outcomes as well. \\

\new{\subsection{Human performance}} 
\new{We perform a human study by asking raters (a disjoint set from the ones who annotated the data) to answer the questions in the dataset. Results are provided in Table \ref{tab:results}. We note that there is a big gap between model and human performance, with the best model (Gemini 2.5 Pro Thinking) still almost 30\% behind human performance. }

\begin{table}[t]
\centering
\small
\begin{tabular}{crr}
\toprule
\textbf{Prompting Method} & \textbf{MCQ Accuracy} & \textbf{\metric{}}\\
\midrule
 Direct Answer & 46.47 & $\dagger$ \\
 + Reasoning & 51.22 & 0.65 \\ 
 + Minerva Rubric & 53.47 & 0.75 \\
\bottomrule
\end{tabular}
\caption{\textbf{Prompting Ablations on \dataset{}.} Results of Gemini 2.0 Flash, 256 frames + ASR. We provide both MCQ accuracy on the final answers and \metric{} on the reasoning traces. $\dagger$ Very few reasoning outputs (by design) to assess.}
\label{tab:prompt-ablation} 
\end{table}

\begin{table}[t]
\centering
\resizebox{0.48\textwidth}{!}{
\small
\begin{tabular}{lrcr}
\toprule
\textbf{Method}  & \textbf{\# Frames} & \textbf{ASR} & \textbf{MCQ-Acc. \%} 
\\
\midrule
Random & - & - & 20.00   \\
\midrule
\textbf{Open-source} \\
Qwen2.5-VL~\cite{bai2025qwen2} & *768 & \cmark & 35.05  \\ 
VideoLLaMA3~\cite{zhang2025videollama} & *180 & \cmark & 35.91 \\
InternVideo2.5~\cite{wang2022internvideo} & 256 & \cmark & 35.18 \\ 
\midrule
\textbf{Proprietary} \\
Claude3.5 Sonnet v2~\cite{claude35sonnetv2} & 64 & \cmark & 31.28\\ 
OpenAI o1~\cite{jaech2024openai}  & 64 & \cmark & 43.48 \\ 
GPT-4o~\citep{achiam2023gpt} & 250 & \cmark & 45.54 \\
\new{GPT-4.1~\cite{openai-gpt4.1}} & \new{256} & \new{\cmark} & \new{53.99} \\
Gemini 2.0 Flash~\citep{reid2024gemini}  & $\dagger$256 & \cmark & 53.47 \\
\new{Gemini 2.5 Flash Thinking~\cite{gemini2.5}} & \new{256} & \new{\cmark} & \new{57.30} \\
\new{Gemini 2.5 Pro Thinking~\cite{gemini2.5}} & \new{1024} & \new{\cmark} & \new{66.20}\\
 
\new{Human performance} & \new{all} & \new{\cmark} & \new{92.54}\\
\bottomrule
\end{tabular}}
\caption{\textbf{Benchmarking performance on \dataset{}.} We report multiple choice accuracy (MCQ-Acc.). * indicates FPS sampling up to frame limit, following optimal settings from \cite{bai2025qwen2,zhang2025videollama}. Since API limits restricted our run of OpenAI o1 and Claude 3.5 Sonnet v2 to 64 frames, we also report results with Flash and GPT-4o with 64 frames in Table \ref{tab:frame_asr_ablations}.$\dagger$ For this model only, we found 256 frames performs slightly better than 2048 (see Appendix).}
\label{tab:results} 
\end{table}

\begin{table}[t]
\centering
\resizebox{0.48\textwidth}{!}{
\small
\begin{tabular}{lrcr}
\toprule
\textbf{Model}  & \textbf{\# Frames} & \textbf{Thinking Off} & \textbf{Thinking On} \\
\toprule 
2.5 Pro & 256 & 63.1 & 64.7 \\ 
2.5 Pro & 512 & 62.3 & 66.0 \\
2.5 Pro & 1024 & 63.9 & \textbf{66.2 }\\
\bottomrule
\end{tabular}}
\caption{\new{\textbf{Thinking ablations on \dataset{} with Gemini 2.5 Pro~\cite{gemini2.5}. MCQ results are provided as Acc.\%.} }}
\label{tab:thinking-results} 
\end{table}

\begin{table*}[ht]
\centering
\resizebox{\textwidth}{!}{
\begin{tabular}{p{4cm}p{2cm}p{10cm}p{12cm}}
\toprule
\textbf{Question} & \textbf{ Answer} & \textbf{Ground truth Reasoning} & \textbf{Model Reasoning} \\
\midrule
In what colors do the aliens speak over the course of the video? & Green and red. & I found 2 aliens at 06:04 and 1 at 08:29. During those times, I examined the text boxes of each speaking alien. I then noticed that the aliens at 06:04 speak in green font, while the alien at 08:29 speaks in red font. Therefore, the colors in which the aliens speak over the course of the video are green and red. & The video content focuses entirely on Scrabble gameplay, discussing various long-word bingos and historical plays. \red{There are no scenes or segments featuring aliens speaking or any reference to colors associated with alien speech. The question appears to be unrelated to the actual content of the video.} Therefore, none of the provided color options (A, B, C, D) are relevant, leading to option E as the correct choice indicating that the premise of the question is incorrect. \\
 
After adding up all the visible numbers on the Wago 221 lever nut when it is first seen in the video what is the total? & 348.34. & At the 05:12 timestamp, the electrician introduces the Wago 221 lever nut into the frame. I noted that 221 was the first number displayed, and I observed as he rotated the device to reveal the side profile view, which featured several numbers. At the 05:14 timestamp, I examined each number, which read as follows: 24, 12, 0.14, 0.2, 4, 5, 50, and 32. I added all the visible numbers to arrive at a sum of 348.34. & The question asks for the sum of all visible numbers on the Wago 221 lever nut when it is first seen in the video. The Wago 221 connector is mentioned in the transcript at timestamp 05:13, \red{but the specific numbers on the connector are not detailed in the text.} However, based on typical Wago 221 models, the numbers usually include '221', '2', and possibly other smaller digits like '1'. Adding these numbers (2 + 2 + 1 + 2 + 1 = 8), and considering potential additional small digits, the closest matching option is (E) 348.34, assuming there may be other minor digits contributing to the total. \\ 
 How many cars can be seen driving by outside while the man in the navy sweater vest first speaks? & 11. & I watched the video to identify the time period that the man in the navy sweater vest first speaks, which is 00:05 to 00:48. From there, I focused my attention on the glass balcony door to his right, where the road and the cars using the road are visible. After that, I counted how many cars passed within the timeframe of 00:05 to 00:48, which came to a total of 11 cars. & Here's my reasoning:

1.  **Identify the relevant time range:** The question asks about cars seen while the man in the navy sweater vest is speaking. He begins speaking around \red{[00:07:15] and continues for a significant portion of the video}.
2.  **Count the cars:** I need to watch the video within the time range identified above and count the cars that pass by outside the window.
3.  **Cars passing by:** I can see 11 cars passing by outside the window.
\\
\bottomrule
\end{tabular}}

\caption{\textbf{Reasoning Failures:} Examples where models produce correct final answers. but have low reasoning scores based on human and \metric{} judgments. Such mistakes often go unnoticed in datasets that only test final answers. Mistakes are highlighted in red.}

\label{tab:error_examples_right_answer} 
\end{table*}
\begin{table}[ht]
\centering
\resizebox{0.45\textwidth}{!}{\begin{tabular}{lcccc}
\toprule

\textbf{Method} & \multicolumn{4}{c}{\textbf{Rubric Criteria: Mean (Pearson \textit{r}) }} \\
& Temporal & Perceptual & Logical & Completeness \\

\midrule

Human  &  0.440 &  0.625 & 0.770 &  0.725 \\
RF-\metric{} & 0.711 (0.56) & 0.684 (0.45) & 0.920 (0.21) & 0.871 (0.07) \\
RB-\metric{}  & 0.434 (0.79) & 0.484 (0.59) & 0.848 (0.17) & 0.748 (0.24) \\

\bottomrule
\end{tabular}}
\caption{\textbf{Humans \& \metric{} for scoring reasoning traces.} We perform an analysis of reasoning traces with our proposed MINERVA Rubric, with both human raters and LLM-based methods (\metric{}; RF: Reference-free, RB: Reference-based), reporting the mean score for each rubric category, and correlation with human judgment (Pearson $r$). We observe that temporally grounded reasoning is a persistent challenge for models.
}
\vspace{-3mm}
\label{tab:metric-dev-set} 
\end{table}

\section{Analysis of Reasoning Traces}  \label{sec:analysis} 

\subsection{Assessing Methods for Video Reasoning}
In this section, we explore methods to score step-by-step rationales generated by multimodal models. Besides increased performance~\cite{wei2022chain,fei2024video}, prompting models to provide reasoning steps also greatly improves interpretability, but objectively studying their correctness and helpfulness (irrespective of the final answer) is still an open problem. While human assessments are usually the gold standard, they are expensive and time-consuming, and so we provide a preliminary exploration of LLM-based assessments as well. 
Within the field of reasoning analysis, a reference-free assessment would be able to score reasoning traces without access to ground truth traces, however would require access to the original video (and can therefore become prohibitively expensive as video length scales). Armed with the traces from \dataset{}, instead we explore \textit{reference-based} analysis, which can operate entirely in the lower-bandwidth (less expensive) text space. We also note that unlike the math or coding domain, where there are multiple ways to achieve the same answer, most video reasoning questions in \dataset{} have less flexibility as they have to be grounded by identifying correct temporal segments and extracting key perceptual information. Our ground truth reasoning traces have the following properties - (i) the number of steps to solve each question is variable, (ii) the reasoning traces are long (Fig.~\ref{fig:dataset-stats}), and (iii) some steps are more important than others. This set-up is challenging for rule-based or embedding-based analysis, but lends itself better to LLM-as-a-judge assessments.
\begin{figure}[t]
\centering
  \includegraphics[width=1\linewidth]{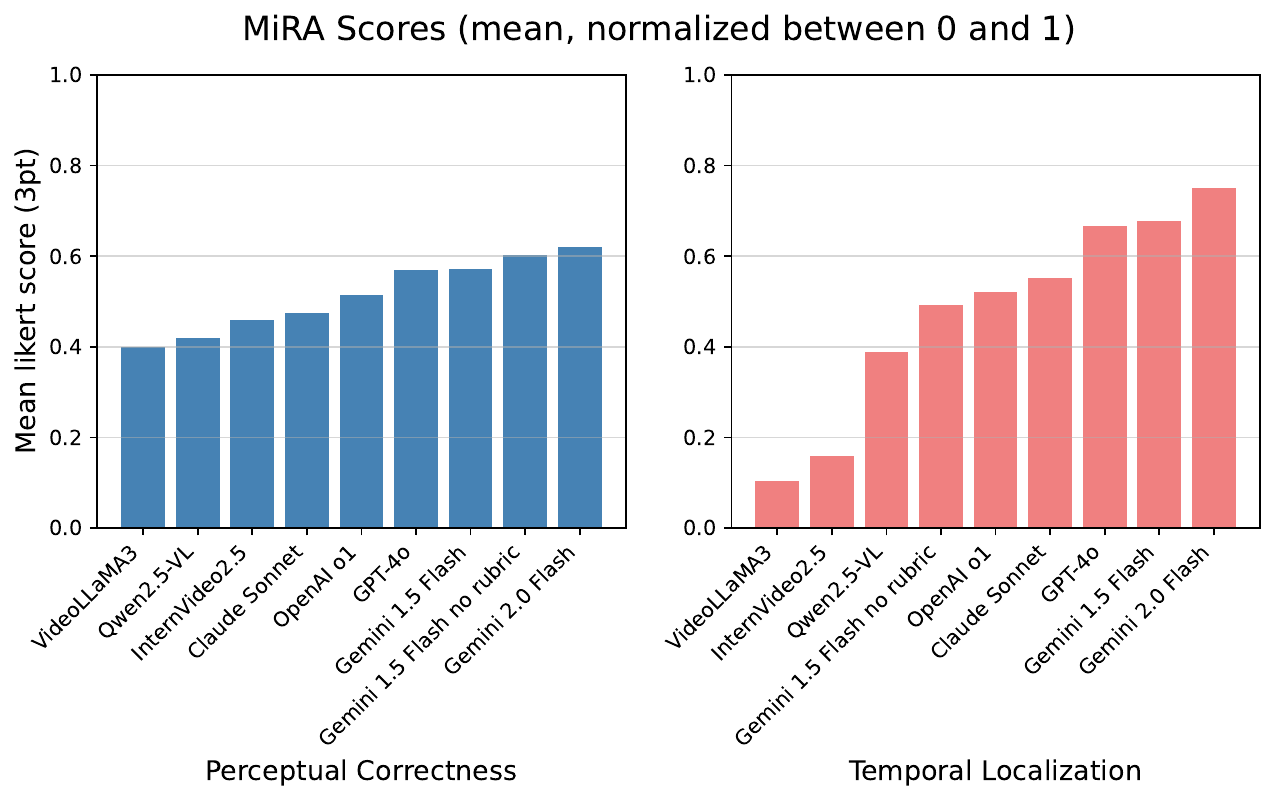}
  \caption{\textbf{Analysis of model produced reasoning traces:} We report \metric{} scores on reasoning traces for two axes in the \dataset{} rubric. Models are presented in ascending order of scores. }
  \label{fig:reasoning-metric-results}
  \end{figure}
We begin by examining model outputs to create a taxonomy of reasoning errors, and then devise a very simple LLM-based judge to score video reasoning. Finally, we collect human evaluation scores on model outputs for a subset of \dataset{}, and analyze model and human correlations. 
\subsubsection{Taxonomy of Video Reasoning Errors}\label{sec:taxonomy}
We propose a simple taxonomy of reasoning errors for video models applied to complex questions. We do this by first examining reasoning outputs from a range of models and identify that errors fall into 4 general categories: \\
\textbf{1. Perceptual Correctness:} This covers errors in perception, including identifying objects, actions, events, or correctly parsing the ASR or performing OCR.  \\
\textbf{2. Temporal Localization:} Errors in identifying the correct temporal range(s) of the video to solve the question. \\
\textbf{3. Logical Reasoning:} Errors in reasoning given the perceptual elements (whether they are correctly identified or not). This includes arithmetic or numerical reasoning errors. \\ 
\textbf{4. Completeness:} A reasoning trace is not complete if it lacks required steps to produce the correct answer.

Examples of errors for each category are provided in Table \ref{tab:error_examples} in the Appendix.
We use these four axes, along with a 3-point Likert score to create a general rubric for evaluating reasoning traces provided by video models, which we call the \dataset{} rubric (we found any more than 3 points to be difficult for both humans and models to provide consistently).
 In the next two sub-sections, we explore human and model assessments for a subset of model traces. We randomly sample 200 questions from \dataset{} and two model outputs for each question from the set \{Gemini 2.0 Flash, GPT-4o, and Qwen2.5-VL\}.
\subsubsection{Human Assessments} \label{sec:metric-dev-set}
 We then provide these 400 model reasoning traces to human raters, along with QADs and ground truth reasoning traces, and ask them to score each reasoning trace with the \dataset{} rubric described above. Note that having the ground truth reasoning traces makes it much quicker for raters, and they do not need to watch a (potentially long) video for each model output. Details of the instructions provided to raters are provided in Appendix. 
\subsubsection{\metric{}: LLM-based Assessments}
We ask an LLM to score reasoning traces according to the \dataset{} rubric -- i.e.\ the same score the human raters provided. We experiment with both reference-based and reference-free prompts. The instructions for human raters to judge model reasoning are provided in Sec.~\ref{appendix:metric-dev-set} in the appendix and the exact prompt for the LLM-as-a-judge is provided in Fig.~\ref{fig:prompt_mira} in the appendix. For clarity, we refer to this assessment henceforth as \metric{} (\dataset{} Reasoning Assessment). Note all scores are normalized to be between 0 and 1. 
\subsection{Results}
\noindent\textbf{Human Scores on Reasoning:} Aggregate human scores for the four axes of the reasoning rubric are shown in Table~\ref{tab:metric-dev-set}, and full distributions are provided in Fig.~\ref{fig:rater_reasoning_score_distributions} (Appendix). We find temporal grounding has the lowest score, followed by perceptual correctness. Scores for logical reasoning and completeness are high, suggesting models are failing largely in the `video' domain, and not text/logical domain. \\
\noindent\textbf{LLM-as-a-judge Scores for Reasoning:} To understand the current limitations of using LLMs as judges for each category in the taxonomy, we examine the correlation between LLM and human judgment  (Table~\ref{tab:metric-dev-set}).
We consider two options for LLM-based automatic ratings: the reference-based (RB) setting, where the model is given the ground truth reasoning annotations as reference with in-context examples for grading with the rubric, and the reference-free (RF) setting, where the model is asked to use the rubric directly.
We use a powerful LLM (Gemini 2.0 Pro) to understand the limits of the best available models in this setting. 
Providing a reference results in substantially stronger correlations along the Temporal, Perceptual, and Completeness axes (particularly for Temporal). It also matches the relative difficulty of each category to human judgment (T < P < C < L). 
Interestingly, having a reference does not necessarily help for the ``Logical'' category -- this appears reasonable as it should focus on \textit{internal} coherence of the model reasoning output (regardless of accuracy, temporal grounding, or length), which depends fully on the LLM's ability to judge novel logic that may dramatically diverge from the ground truth reference.
Thus, our analysis reveals new avenues for improvement for using LLM-based judges for video reasoning evaluation.
\\
\noindent\textbf{Scoring reasoning using (\metric{}) on the full set of questions:} 
The advantage of our references is that we can scale the results for the reference-based LLM evaluation for \new{multiple models (here we choose 8 of the models, covering both open-source and proprietary models)} to the full dataset with limited manual effort. We show scores for the Temporal and Perceptual categories, where LLMs yield reliable representations of human judgments in Fig.~\ref{fig:reasoning-metric-results}.
Raw scores and plots for the other two categories are in Fig.~\ref{fig-appendix:reasoning-metric-results} in the appendix for completeness.
Interestingly, our preliminary analysis shows gaps between the 3 open-source models (which perform similarly on MCQ eval) -- with Qwen2.5-VL far outperforming the others on temporal localization (see qualitative examples in Tab.~\ref{tab:reasoning_qualitative_comparison} in the appendix). 
We also find scores are lowest for temporal localization and perceptual correctness, with the largest variation among models in the former.
This mirrors the results from the human study, suggesting that when video models with heavy, well-trained LLMs are applied to video reasoning, they tend to provide plausible sounding detailed reasoning traces, however they still lack some key perceptual insights and temporal grounding in the video. 

\section{Conclusion}
We present \dataset{}, a videoQA dataset with complex questions, answers and ground truth reasoning traces. Our dataset is challenging for multiple frontier multimodal models, and is useful for providing insights into the reasoning failures of these models. Our analysis of using LLMs to judge model-generated reasoning traces shows promise and points out opportunities for future work in this direction.

\new{\section{Acknowledgements}}
\new{We are grateful to Antoine Yang, Anelia Angelova, Mario Lucic, Karel Lenc and Boyu Wang from the Gemini team for their support.}
\clearpage
{
    \small
    \bibliographystyle{ieeenat_fullname}
    \bibliography{main}
}

\clearpage
\setcounter{page}{1}
\maketitlesupplementary

\section{Rater Guidelines}
All textual data in MINERVA was entirely created by human annotators (raters). All raters are native English speakers with graduate degrees.
Here we provide guidelines given to the raters for dataset creation (Sec. \ref{appendix:dataset_creation_guidelines}), 
and for scoring reasoning (Sec. \ref{appendix:metric-dev-set}).  
\subsection{Dataset Creation}
\label{appendix:dataset_creation_guidelines}
The raters were given the following guidelines before being asked to propose question, answers, decoys and reasoning traces. 

\subsubsection{Good Questions} \label{appendix:question-raters}
\begin{itemize}
    \item{The question should not be easily solvable by looking at just a few frames in the video} 
\item{It should not be solvable using only common sense and external knowledge}
\item{It should ask about visual elements in the video (and not just focus on the speech)}
\item{It should not be subjective and should have only one right answer}
\item{It should be complex, and require multiple steps to solve}
\item{It should not be offensive}
\item{Please do not mention any names of humans in the question or the answer unless they are fictional, famous or celebrities}
\item{Each question should belong to at least two question types}
\end{itemize}
\subsubsection{Reasoning Traces} 
\begin{itemize}
\item{A reasoning step is an action that you would take to break down the question solving process. You can think of them as the building blocks to the solution.}
\item{A good question requires multiple reasoning steps to be performed in sequence to arrive at the answer.}
\item{Without one of the steps, a person should not be able to get the answer.}
\item{Do not add irrelevant information in the steps.}
\item{The final answer can be very short eg. a single word. However, the reasoning steps are the entire process to get to the answer.}
\end{itemize}
\subsubsection{Decoys} 
\begin{itemize}
    \item A decoy is a wrong answer to the question. We need decoys to create multiple choice questions, like in a multiple choice exam. For our questions, we want to provide 5 options where only one is correct. 
\item Decoys should be diverse. They should be different enough from each other to not narrow down the scope of the question too much.
\item The correct answer should not stand out among the decoys. So, decoys should not have obvious differences to the answer. For example,
\begin{enumerate}
    \item Decoys should not be longer or shorter than the true answer.
    \item A decoy should not be an impossible/implausible answer.
    \item A decoy should not make the task easy to solve without watching the video. 
\end{enumerate}
\end{itemize}
\subsubsection{Pipelining and Quality Control} \label{appendix:pipelining}
The flow is as follows: \\
\noindent\textbf{Initial Annotation (Curator):} A rater familliar with the subject matter of the video domain (eg for a basketball video,  are asked to ensure they know the rules) watches the video and creates 5 distinct questions. This is done concurrently by two raters, to get 10 questions at the end. Each rater works independently to ensure uniqueness and skill diversity. \\
\noindent\textbf{Peer Review:} The initial annotations, including the questions, are then passed to another rater for peer review. This reviewer checks for question complexity and suggests corrections or improvements. (this was implemented during early curation phases and was eliminated later as the raters got more proficient). \\
\noindent\textbf{Senior Review:} Following the peer review, a Senior Reviewer examines the work. The Senior Reviewer makes any necessary final changes to ensure quality and adherence to the established standards.

\new{\subsection{Human Study}} \label{appendix:human_study_guidelines}
\new{Goal: You will be given a video, a question, and 5 answer choices. Please watch the video and pick the correct answer. (Like a multiple choice exam). You can watch the video as many times as you like, and you can rewatch various parts of the video. Please take as much time as you require.} \\
\new{\noindent\textbf{Pipelining and Quality Control}: We ensure that the raters for dataset creation and the human study are disjoint. 10\% of the data is checked by another rater to gauge interrater agreement for the human study.} \\
\subsection{Human Assessments of Reasoning}\label{appendix:metric-dev-set}
\textbf{Goal:} You will be given a video and a question. You will also see the outputs of two models which have been asked to answer the question and provide their reasoning. \\
Your task is to provide scores for the reasoning. \\
You can judge the reasoning based on the following criteria: \\
(1) Perceptual correctness: was the relevant information perceived accurately from the video? (eg were the correct objects identified, was the text read properly from the screen, were the relevant events and actions mentioned correctly) \\
(2) Temporal grounding: were time ranges provided for each piece of information from the video, and if so were they accurate? \\
(3) Logical reasoning: was the reasoning logically sound, given the information perceived (independent of whether that information was correct)? \\
(4) Completeness: were any steps skipped in the given answer or left unstated? \\

\noindent{For each of the above criteria, please provide a score from 0,1,2.} \\
0 - Doesn’t fulfill the criteria at all. \\
1 - Partially fulfills the criteria. \\
2 - Completely fulfills the criteria. \\

\section{Model Baselines}
\subsection{Prompts} \label{appendix:model-prompts}

\begin{figure}[ht]
    \centering
    \small
\begin{exampleprompt}
    \texttt{You will be given a question about a video and five possible answer options. You are provided frames from the video, sampled evenly across the video.\\ \\
    Transcript: \{asr\} \\
    Frames: \{frame1\}, \dots, \{frame $N$\}\\
    Question: \{question\}\\
    Possible answer choices: \{answer choices\} \\
    \\
    Output the final answer in the format ``Final Answer: (X)'' where X is the correct digit choice. DO NOT OUTPUT text or any other words with the full answer.
    }
\end{exampleprompt}
    \vspace{-\baselineskip}
    \caption{\textbf{Direct MCQ prompt for Gemini.}}
    \vspace{-\baselineskip}
    \label{fig:prompt_mcq_answer_direct}
\end{figure}

\begin{figure}[ht]
    \centering
    \small
\begin{exampleprompt}
    \texttt{You will be given a question about a video and five possible answer options. You are provided frames from the video, sampled evenly across the video.\\ \\
    Transcript: \{asr\} \\
    Frames: \{frame1\}, \dots, \{frame $N$\}\\
    Question: \{question\}\\
    Possible answer choices: \{answer choices\} \\
    \\
    After explaining your reasoning, output the final answer in the format ``Final Answer: (X)'' where X is the correct digit choice. Never say ``unknown'' or ``unsure'', or ``None'', instead provide your most likely guess.
    }
\end{exampleprompt}
    \vspace{-\baselineskip}
    \caption{\textbf{Reasoning MCQ prompt for Gemini.}}
    \vspace{-\baselineskip}
    \label{fig:prompt_mcq_answer_reasoning}
\end{figure}

\begin{figure}[ht]
    \centering
    \small
\begin{exampleprompt}
    \texttt{You will be given a question about a video and five possible answer options. You are provided frames from the video, sampled evenly across the video.\\ \\
    Transcript: \{asr\} \\
    Frames: \{frame1\}, \dots, \{frame $N$\}\\
    Question: \{question\}\\
    Possible answer choices: \{answer choices\} \\
    \\
    Provide all steps required to come to the answer in your reasoning, and the following rubric will be used to judge the reasoning:\\
    (1) Perceptual correctness: was the relevant information perceived accurately from the video?\\
    (2) Temporal grounding: were time ranges provided for each piece of information from the video, and if so were they accurate? \\
    (3) Logical reasoning: was the reasoning logically sound, given the information perceived (independent of whether that information was correct)? \\
    (4) Completeness: were any steps skipped in the given answer or left unstated? \\ \\
    After explaining your reasoning, output the final answer in the format ``Final Answer: (X)'' where X is the correct digit choice. Never say ``unknown'' or ``unsure'', or ``None'', instead provide your most likely guess.
    }
\end{exampleprompt}
    \vspace{-\baselineskip}
    \caption{\textbf{Reasoning MCQ prompt for Gemini with the Minerva Rubric.}}
    \vspace{-\baselineskip}
    \label{fig:prompt_mcq_answer_rubric}
\end{figure}

\begin{figure*}[ht]
    \centering
    \tiny
\begin{exampleprompt}
\texttt{You are an expert at grading student answers to questions about videos. For each video, you will get a question about the video, the correct reasoning, and the final answer. You will then get the reasoning from the student, and a set of criteria. Given this criteria, please provide a score from 0, 1 or 2 for each criterion that will assess the student's work.}

\begin{verbatim}
**Criteria:**
(1) Perceptual correctness: was the relevant information perceived accurately from the video?
(2) Temporal grounding: were time ranges provided for each piece of information from the video,
and if so were they accurate?
(3) Logical reasoning: was the reasoning logically sound, given the information perceived (independent
of whether that information was correct)?
(4) Completeness: were any steps skipped in the given answer or left unstated?

For each of the above criteria, please provide a score from 0,1,2.
0 - Doesn’t fulfill the criteria at all
1 - Partially fulfills the criteria
2 - Completely fulfills the criteria
Please produce the score in the JSON format:
```
{"Perceptual correctness": <score that is 0,1,2>, "Temporal grounding": <score that is 0,1,2>, "Logical reasoning": <score that is 0,1,2>, 
"Completeness": <score that is 0,1,2>}
```

**Examples:**

**Question:** What has to happen, according to the performer, for the wishing audience member's wish
to come true? The wishing audience member must stand up., The wishing audience member must come up 
onstage., The entire audience must close their eyes., The entire audience must imagine the wishing 
audience member naked., The wishing audience member must receive truth.,
**Reference Answer:** I watched the sequence in the video where the performer dresses up at the 
"Magic Magic Wish Man" from 03:18 to 04:00, and singles out an audience member to make a wish at 
03:38. I then listened for the performer to tell the audience how to make the wish come true, and
from 03:50 to 03:55, he tells the entire audience that they must close their eyes.
**Student Reasoning** The entire audience must close their eyes.

Output should be:
```
{"Perceptual correctness": 2, "Temporal grounding": 0, "Logical reasoning": 2, "Completeness": 0}
```

**Question:** How many hearts are visible in the picture of the finished building at 00:37?8., 10.,
7., 11., 9.,
**Reference Answer:** I watched the video until the indicated time code of 00:37. From there, I
counted the number of red and blue hearts visible in the photo: 4 on the columns, 1 on the entrance 
roof, 1 on the left edge of the building, and 3 on the top facade. This comes to a total of 9 visible
hearts in the photo.
**Student Reasoning** There are five hearts visible on the building in the image at 00:37.

Output should be:
```
{"Perceptual correctness": 1, "Temporal grounding": 2, "Logical reasoning": 0, "Completeness": 1}
```

**Question:** How did the woman in green react to the woman in blue crying? She became worried., She cried with her., She became annoyed., She became indifferent.,
She showed empathy.,
**Reference Answer:** I watched the video until 1:58, where the woman in blue cried. Afterward, the woman in green asked her if she plays video games. The woman in 
blue responded by saying she plays SIMS and pretends that they're her family. Then, the woman in green looked at the man in red and widened her eyes, a clear sign 
that she was worried.
**Student Reasoning** At around 2 minutes into the video, there's a scene where the woman in blue starts crying while sitting on the couch. The woman in green, who 
is seated next to her, leans forward, places a hand on her shoulder, and speaks to her in a comforting manner. This action clearly indicates that she is showing 
empathy towards the woman in blue. There is no indication of worry, annoyance, indifference, or crying along, which rules out options A, B, C, and D.

Output should be:
```
{"Perceptual correctness": 0, "Temporal grounding": 1, "Logical reasoning": 1, "Completeness": 0}
```

**Question:** How many passes did the white team make right before they scored their first goal? 4., 1., 2., 5., 3.,
**Reference Answer:** I searched for the white team's first score, which happened at 00:42. I watched the clip that showed the entire play, which is from 00:31 to 
00:43. I counted one pass at 00:39 and one pass at 00:41. The player who receives the pass at 00:41 shoots into the goal. To confirm, I watched the replay from 
00:44 to 00:47. At 00:46, the puck bounces off a white player's stick and into the goal. This makes the shot taken at 00:41 a pass and assist instead of a proper 
shot, which means the white team made 3 passes before they scored their first goal.
**Student Reasoning** In the frames leading up to the first goal scored by the team in white, we can see a sequence where the puck is passed between players. The 
sequence goes as follows:\n\n1. A white team player makes a pass to a teammate in the offensive zone.\n2. The teammate receives the puck and makes a quick pass to 
another player.\n3. The third player receives the pass and immediately shoots on goal.\n\nThis sequence involves a total of three passes made by the white team 
right before the goal is scored.\n\nFinal Answer: (E)

Output should be:
```
{"Perceptual correctness": 2, "Temporal grounding": 0, "Logical reasoning": 2, "Completeness": 2}
```

**Re-iterating the Criteria**
(1) Perceptual correctness: was the relevant information perceived accurately from the video? (eg were the correct objects identified, was the text read properly 
from the screen, were the relevant events and actions mentioned correctly)
(2) Temporal grounding: were time ranges provided for each piece of information from the video, and if so were they accurate?
(3) Logical reasoning: was the reasoning logically sound, given the information perceived (independent of whether that information was correct)?
(4) Completeness: were any steps skipped in the given answer or left unstated?

**Final Instruction and Input**
Please now produce the scores (0, 1, or 2) in the correct JSON format for the following question, reference answer, and student reasoning, following the criteria.
**Question:** {question} {formatted_answer_options}

**Reference Answer:** {reference_reasoning} Final Answer: {reference_final_answer}

**Student Reasoning:** {model_reasoning}

JSON output and justification:
\end{verbatim}
\end{exampleprompt}
    \vspace{-\baselineskip}
    \caption{\textbf{Scoring prompt for MiRA analysis.} We show the main ``reference-based'' prompt above, which includes reference examples with ground-truth reference reasoning traces, as well as the reference ground-truth reasoning trace for the question + reasoning being evaluated. The ``reference-free'' version run for comparison in the main paper omits this. In both prompts we provide our MINERVA rubric.}
    \vspace{-\baselineskip}
    \label{fig:prompt_mira}
\end{figure*}

Section~\ref{sec:promptsection} presents an ablation study on various prompting strategies. 
Here, we provide details on each strategy. 
Specifically, Figure~\ref{fig:prompt_mcq_answer_direct} illustrates the Direct MCQ prompt, Figure~\ref{fig:prompt_mcq_answer_reasoning} shows the reasoning prompt, and Figure~\ref{fig:prompt_mcq_answer_rubric} displays the reasoning prompt incorporating the Minerva Rubric.

\subsection{Implementation Details}
\subsubsection{Hyperparameters}
Hyperparameters for all our models are provided in Table \ref{tab:hyperparams}. 
\begin{table*}[ht]
\centering
\begin{tabular}{lllr}
\toprule
\textbf{Method}  & \textbf{\# of Frames} & \textbf{ASR} & \textbf{Hyperparameters (seeds, temperature, etc)}  \\
\midrule
InternVideo2.5~\cite{wang2022internvideo} & 256 & \cmark & \parbox{6cm}{image size=448, \\ temperature=0,\\ top-p=0.1 (default), \\ beams=1, \\sample=False,}\\ 
\midrule
Qwen2.5-VL-72B~\cite{bai2025qwen2} & 768* & \cmark & \parbox{6cm}{frames=2fps up to 768 frames (default),\\ seed=default,\\ sampling=default} \\ 
 
\midrule
VideoLLaMA3-7B~\cite{zhang2025videollama} & 180* & \cmark &  \parbox{6cm}{frames=1fps up to 180 frames (default),\\ seed=default,\\ sampling=default}\\
\midrule
Deepseek-R1:32b & 0 (blind) & \cmark & \parbox{6cm}{seed=default, \\ temperature=1 (default),\\top-p=default,\\}\\ 
\midrule
GPT-4o~\citep{achiam2023gpt} & 250 & \cmark &  \parbox{6cm}{version=gpt-4o-2024-08-06, \\ seed=default, \\ top-p=default, \\ temperature=1 (default), \\ image resolution model=low} \\
\midrule
\new{GPT-4.1~\cite{openai-gpt4.1}} & 256 & \cmark &  \parbox{6cm}{version=gpt-4o-2024-08-06, \\ seed=default, \\ top-p=default, \\ temperature=1 (default), \\ image resolution model=low} \\
\midrule
Claude 3.5 Sonnet v2~\cite{claude35sonnetv2} & 64  & \cmark & \parbox{6cm}{image\_size=448, \\ other parameters = default } \\ 
\midrule
Gemini 1.5 Pro~\citep{reid2024gemini} & all & \cmark  & \parbox{6cm}{temperature=0, \\ seed=default, \\ sampling=default}  \\
\midrule
Gemini 2.0 Flash~\citep{reid2024gemini}  & all & \cmark & \parbox{6cm}{temperature=0, \\ seed=default, \\ sampling=default} \\
\midrule
\new{Gemini 2.5 Flash Thinking~\citep{gemini2.5}}  & all & \cmark & \parbox{6cm}{temperature=0, \\ seed=default, \\ sampling=default} \\
\midrule
\new{Gemini 2.5 Pro Thinking~\citep{gemini2.5}}  & all & \cmark & \parbox{6cm}{temperature=0, \\ seed=default, \\ sampling=default} \\
\midrule
OpenAI o1~\cite{jaech2024openai} & 64 & \cmark & \parbox{6cm}{image\_size=448, \\ reasoning\_effort=medium, \\ other parameters = default}\\  
\bottomrule
\end{tabular}
\caption{\textbf{Hyperparameters for all model baselines}}
\label{tab:hyperparams} 
\end{table*}
\subsubsection{API Access Dates}
We accessed each API for all API-based models between Feb 24 and Mar 7, 2025.
\section{Ablations}
\begin{figure}[ht]
\centering
  \includegraphics[width=0.9\linewidth]{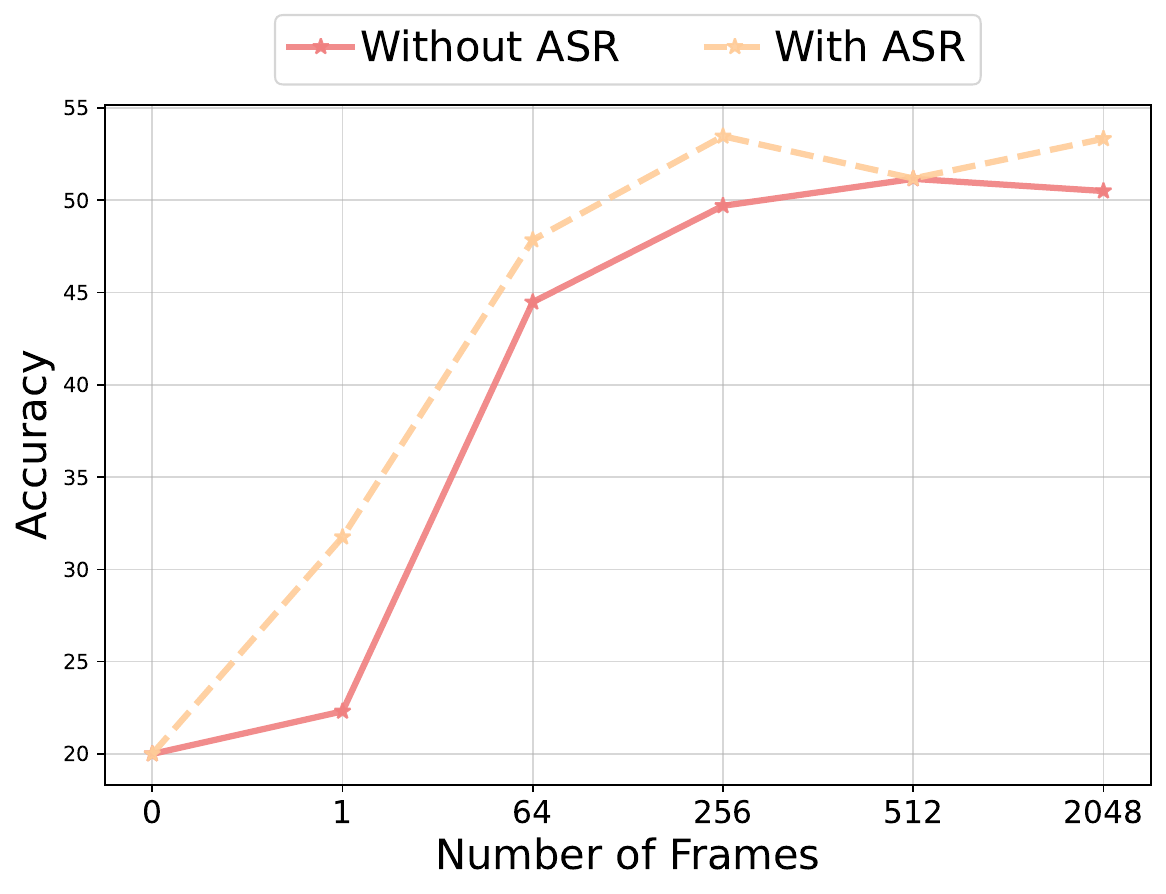}
  \caption{\textbf{Results with varying number of frames using Gemini 2 Flash:} Results appear to saturate around 256 frames (note: x-axis is not linear scale).}
  \label{fig:appendix-frame-ablations}
  \end{figure}

We show a fine-grained frame ablation for Gemini 2.0 Flash in Fig. \ref{fig:appendix-frame-ablations}. Results appear to saturate around 256 frames for this model, which also sets the state-of-the-art on \dataset{}.
\section{Statistics for \dataset{}}
\begin{figure*}[ht]
\centering
  \includegraphics[width=0.9\linewidth]{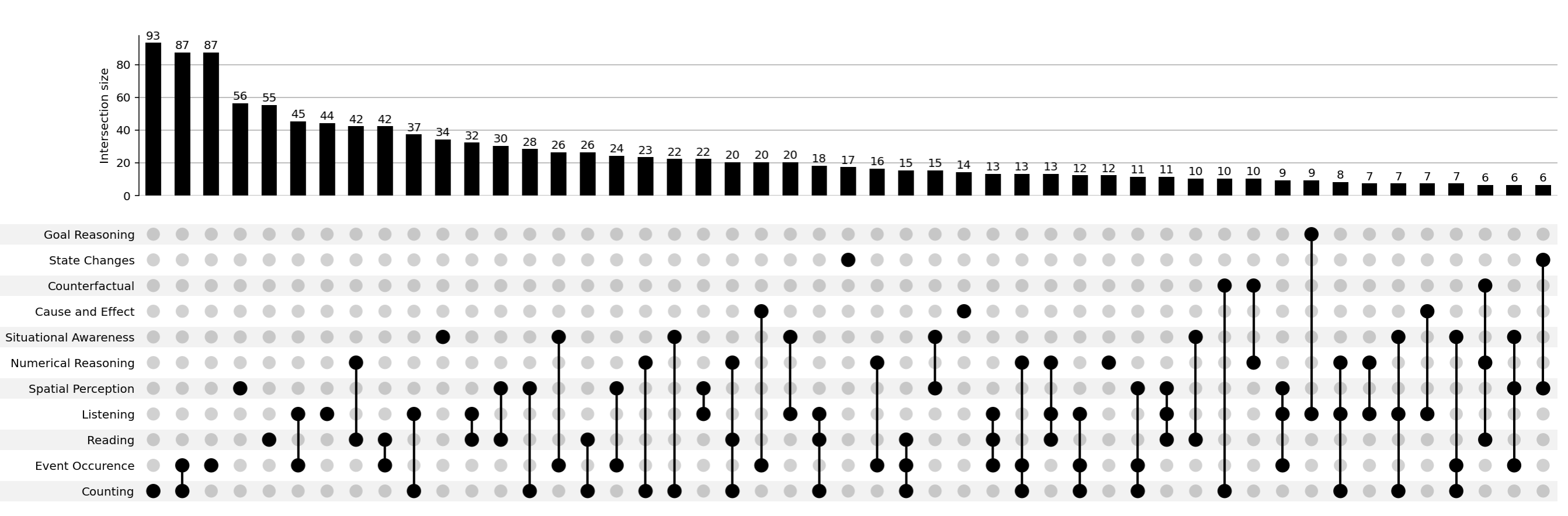}
  \caption{\textbf{Distribution of skill types:} Given each question requires two or more skills, we show an upset plot of combinations of skills. We omit Object Recognition and Temporal Reasoning, as they are required for almost all questions.}
  \label{fig:skills-upset}
  \end{figure*}

We provide a distribution of skill types for the questions in the dataset in Fig. \ref{fig:skills-upset}. Note each question in the dataset requires two or more skills, and hence we show an upset plot of combinations of skills. We omit Object Recognition and Temporal Reasoning, as they are required for almost all questions.
\section{Reasoning Analysis}
\begin{figure*}[ht]
\centering
  \includegraphics[width=0.9\linewidth]{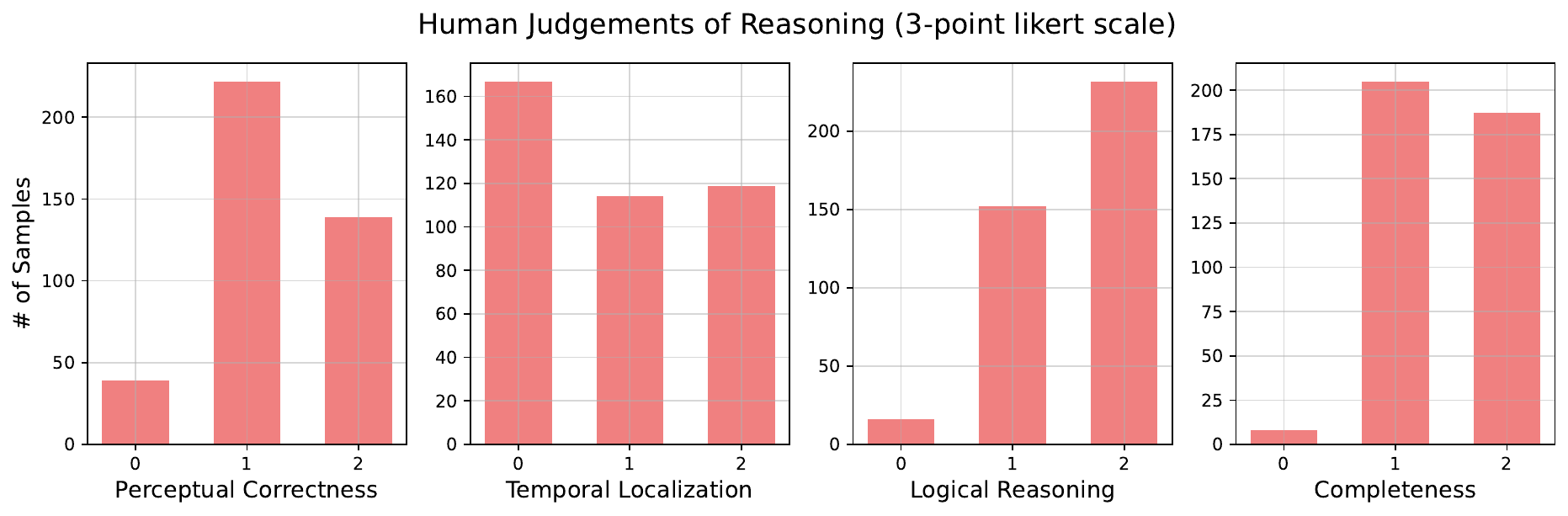}
  
  \caption{\textbf{Distribution of raters scores on 400 model reasoning traces for 200 questions:} Lowest scores are for temporal localization, followed by perceptual correctness. 
  }
  \label{fig:rater_reasoning_score_distributions}
  \end{figure*}

\begin{figure*}[t]
\centering
  \includegraphics[width=0.9\linewidth]{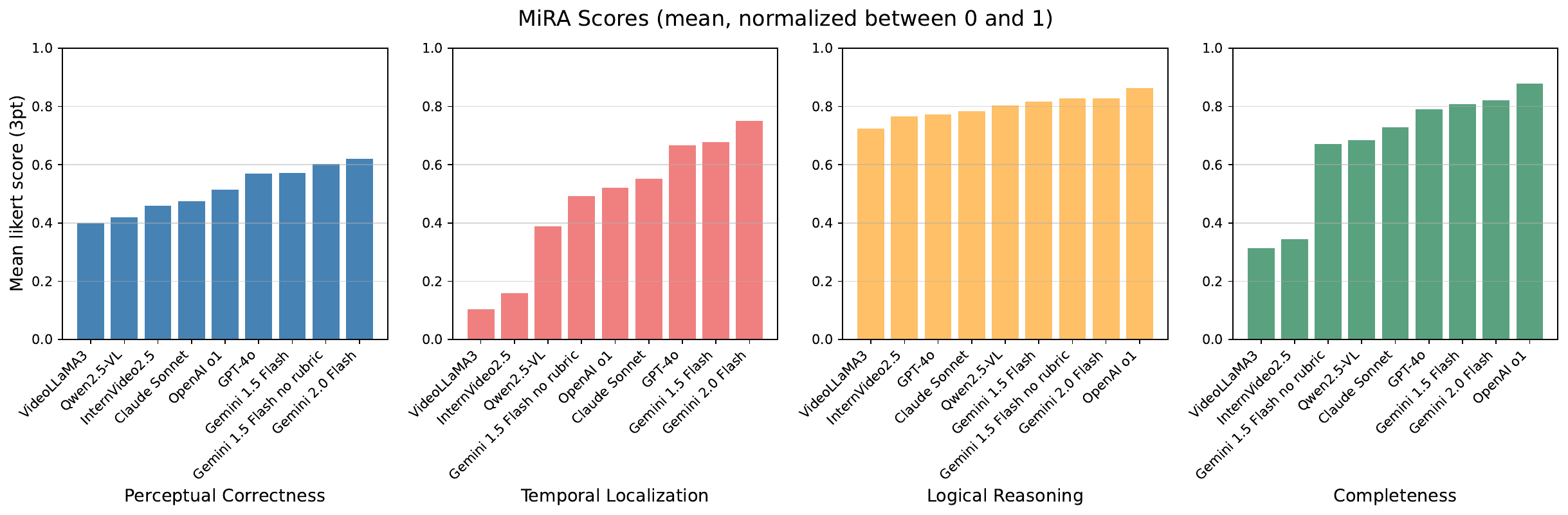}
  \caption{\textbf{Analysis of model produced reasoning traces on \dataset{} with \metric{}:} We report \metric{} scores on reasoning traces for all 4 different axes in the \dataset{} rubric. Models are presented in ascending order of scores. Results are lower for perceptual correctness and temporal grounding than logical reasoning and completeness. This suggests that when asking frontier multimodal models (which contain heavy, well-trained LLMs) reasoning-heavy video questions, they tend to provide plausible sounding detailed reasoning traces, however they lack key perceptual insights and temporal grounding in the video. }
  \label{fig-appendix:reasoning-metric-results}
  \end{figure*}

Examples of the \dataset{} taxonomy are provided in Table. \ref{tab:error_examples}. We provide the distribution of rater scores on the 400 examples chosen from \dataset{} in Fig. \ref{fig:rater_reasoning_score_distributions}, and the scoring prompts used for the \metric{} analysis in Figure \ref{fig:prompt_mira}. Results showing \metric{} scores for all 4 axes of the rubric on the full dataset can be found in Fig. \ref{fig-appendix:reasoning-metric-results} and Table \ref{appendix:results}. Table~\ref{tab:reasoning_qualitative_comparison} shows how the quality of reasoning traces can differ dramatically even though the models achieved comparable MCQ performance.


\begin{table*}[ht]
\centering
\resizebox{0.9\textwidth}{!}{
\begin{tabular}{lllrrrrrr}
\toprule
\textbf{Method}  & \textbf{\# of Frames} & \textbf{ASR} & \textbf{MCQ-Acc. \%} & \multicolumn{5}{c}{\textbf{\metric}} \\ 
& & & & P & T & L & C & Total \\ 
\midrule
Random & - & - & 20.00   \\
\midrule
\textbf{Open-source} \\
Qwen2.5-VL~\cite{bai2025qwen2} & 768* & \cmark & 35.05 & 0.42 & 0.39 & 0.80 & 0.68 & 0.57\\ 
VideoLLaMA3~\cite{zhang2025videollama} & 180* & \cmark & 35.91 & 0.40 & 0.10 & 0.72 & 0.31 & 0.39 \\
InternVideo2.5 & 256 & \cmark & 35.18 & 0.46 & 0.16 & 0.77 & 0.34 & 0.43\\ 
\midrule
\textbf{Proprietary} \\
Claude3.5 Sonnet v2~\cite{claude35sonnetv2} & 64 & \cmark & 31.28 & 0.47 & 0.55 & 0.78 & 0.73 & 0.63 \\ 
OpenAI o1~\cite{jaech2024openai}  & 64 & \cmark & 43.48 & 0.52 & 0.52 & \textbf{0.86} & \textbf{0.88} & 0.69 \\ 
GPT-4o~\citep{achiam2023gpt} & 250 & \cmark & 45.54 & 0.57 & 0.67 & 0.77 & 0.79 & 0.70 \\
Gemini 1.5 Flash~\citep{reid2024gemini} & 2048 & \cmark  &  49.31 & 0.57 & 0.68 & 0.82 & 0.81 & 0.72 \\
Gemini 2.0 Flash~\citep{reid2024gemini}  & 256 & \cmark & \textbf{53.47} & \textbf{0.62} & \textbf{0.75} & 0.83 & 0.82 & \textbf{0.75} \\
\bottomrule
\end{tabular}}
\caption{\textbf{Benchmarking performance on \dataset{}.} We report multiple choice accuracy (MCQ-Acc.) and \metric scores normalized to be between 0 and 1. P: Perceptual Correctness, T: Temporal Localization: L: Logical Reasoning: C: Correctness * indicates FPS sampling up to frame limit, following optimal settings from prior work\cite{bai2025qwen2,zhang2025videollama}.}

\label{appendix:results} 
\end{table*}
\begin{table*}[ht]
\centering
\resizebox{\textwidth}{!}{
\begin{tabular}{cp{4cm}p{2cm}p{9cm}p{8cm}}
\toprule
\textbf{Error Category} & \textbf{Question} & \textbf{ Answer} & \textbf{Ground truth Reasoning} & \textbf{Model Reasoning} \\
\midrule

\textbf{Temporal Localization} & What happens after the announcer announces the reward for winning the match? & The crowd loudly boos. & At 02:15 - 02:21, I heard the announcer announce "Whoever wins out of you two enters the rumble last". Immediately after, I then heard the surrounding crowd loudly boo. Therefore, the crowd booed after the announcer announced the reward for winning the match. & The announcer explains that the winner of the match will enter the rumble last, and the loser will enter first.  Immediately following this announcement \red{(around 03:04-03:06)}, the video shows the two wrestlers beginning to fight. \\ 
\midrule
\textbf{Perceptual Correctness} & What is the significance of Aja Wilson's last shot attempt? & The game would have been tied if it went in. & At 00:30-00:40, I heard the announcer say that Aja Wilson was dominant all game for the Gamecocks as I saw a player make a shot. At 00:37, the player turned and walked away and I saw the number "22" and the word "Wilson" written on her back, indicating that Aja Wilson wears the number 22. Then, I watched this player's last shot attempt from 01:47-01:52. The shot was taken close to the basket and inside the 3-point line, so it is a 2-point attempt. At 01:53, I saw the final score, which would've been 62-62 if Aja Wilson had made her last shot. Therefore, the game would have been tied if the shot went in. & The video shows that the \red{game was tied at 60-60 } when Aja Wilson attempted her last shot.  The score shown on the screen at 1:23 is \red{60-60.}  If Wilson's shot had been successful, South Carolina would have taken the lead and likely won the game, as there was little time remaining. \\ 
\midrule 
\textbf{Logical Reasoning} & What is the mans name that is sitting down inside of the cage? & Jacob. & I watched as the two characters from the beginning of the video enter the doorway. At 00:39, one of the characters yells "Jacob" as he runs towards the cage. Inside the cage we can see a man sitting, at 00:41, who responds to the name "Jacob". & The man inside the cage is \red{referred to by the name 'Ryan'} when another character says, 'Hey! Hey, it's Ryan! Jesus, what happened to you..?' This dialogue occurs around the 45-second mark of the video.\\ 
\midrule
\textbf{Completeness} & If each settlement is surrounded by 3 tiles with numbers, what is the sum of the numbers on the tiles surrounding both the blue and black settlements at 05:10? & 47. & I found the blue and black settlements on the board at 05:16. At the same time, I noticed the blue settlement was surrounded by 3 tiles with the numbers "5", "8", and "10". Likewise, I saw the black settlement was surrounded by 3 tiles with the numbers "9", "11", and "4". I then calculated 5+8+10+9+11+4=47. Therefore, the sum of the numbers on the tiles surrounding both the blue and black settlements at 05:10 is 47. & The sum of the tiles gives a total of 49. Therefore, the correct answer is 49. \\ 

\bottomrule
\end{tabular}}
\caption{\textbf{Taxonomy of Video Reasoning Errors.} Examples are provided from \dataset{}, and the model used is Qwen2.5-VL-72B. Note how in the first example, the model fails to correctly perceive the score to be 62-60 at the end, which leads to the wrong conclusion, while in the second example the model incorrectly localizes the announcement to start at 03:04, instead of 02:15. In the fourth example, the model concludes that the man in the cage is called Ryan simply because another character uses the name Ryan, which is not correct. In the final example, the model simply produces the final answer without showing any steps. }
\label{tab:error_examples} 
\end{table*}

\begin{table*}[ht]
\centering
\resizebox{\textwidth}{!}{
\begin{tabular}{p{4cm}p{2cm}p{8cm}p{8cm}p{8cm}}
\toprule
\textbf{Question} & \textbf{ Answer} & \textbf{Ground truth Reasoning} & \textbf{Qwen2.5-VL Reasoning} & \textbf{InternVideo2.5 Reasoning} \\
\midrule
What is the unique advantage Player 32 from Bristol Central has over the opposing team? &
He is the tallest player among both teams. &
I looked at the video in its entirety to locate instances where Player 32 from Bristol Central scored points or blocked the opponents. At 00:21 he makes his first block, at 00:33 he scores his first point, at 00:54 and 01:01 he makes 2 more blocks, and at 01:30 he makes his first dunk. From looking at high angles shots of the basketball court, such as the one at 01:08, it is evident that Player 32 is the tallest player among both teams. This information helped me to conclude that Player 32's unique advantage is his height. &
Throughout the video, Player 32 from Bristol Central is consistently seen dominating the paint area due to his height advantage. This is evident when he successfully grabs rebounds and scores over the heads of the opposing players. The visual evidence shows that he towers over other players during key moments like the 1:43 mark where he secures a rebound despite being surrounded by defenders. His height allows him to reach higher for shots and blocks, which is a significant advantage in basketball. &
Player 32 from Bristol Central is consistently shown as the tallest player on the court, making him a dominant presence in the paint. This advantage allows him to score more points and outplay the opposing team's players. \\
\midrule
What does the basketball coach do after the interviewer asks his first question? &
He scratches his head. &
I looked for the interviewer's first question. This happens at the beginning of the video, from 00:00, and it goes to 00:04. At this point, the interviewer finishes his question and stops talking. At this point, I watch the basketball coach to see his response. As he responds, he raises a hand to the side of his head and scratches. &
After the interviewer asks the first question, the basketball coach can be seen scratching his head while responding. This gesture occurs shortly after the initial question is posed, indicating a moment of reflection or emphasis as he begins to elaborate on the team's performance. The visual evidence from the video clearly shows this action, aligning with option (C). &
After the interviewer asks his first question, the basketball coach is seen scratching his head. \\
\midrule
How many shots on goal were taken that were saved by the goalie of either team, not including replays? &
6. &
I watched the video for all shots taken on a goal. After discarding the shots that scored, the video shows ten shots taken that were saved: 00:16, 00:17, 00:23, 00:25, 00:27, 01:01, 01:37, 01:40, 01:45, 02:05. I noticed that four of these were replays: 00:17, 00:27, 01:40, 01:45. After subtracting replays from shots taken and saved, the answer is (1) Six. &
Based on the video and the provided transcript, we can identify several instances where shots were taken and saved by the goalies. At around 00:25, a shot was deflected off the crossbar, indicating a save. Around 00:47, another shot was mentioned but not scored, suggesting a save. At 01:15, a shot was described as being saved. Additionally, at 01:58, a shot was blocked by a stick. Finally, at 02:09, a shot was let go but not scored. These five saves, along with one more visible in the video near the end of the first period, total six saves. Therefore, the correct answer is E. &
The video shows several shots on goal that were saved by the goalie of either team. By counting these shots, we can determine the answer. \\
\midrule
What two cards, including their suits, are Hawk Tuah holding as part of her 2-pair hand in hand three? &
King of Hearts, Queen of Spades. &
I listened to the narrator introduce hand 3 at 04:28. Just after that time, \textbf{at 04:30}, I watched as Hawk Tuah's hand was displayed on the left side of the screen, to show a King of Hearts and Queen of Spades just above her name. &
In the video, during the discussion of Hand Three, the on-screen graphics clearly show that Hawk Tuah is holding the King of Hearts and Queen of Spades. This can be seen around the timestamp where the flop is revealed and the players' hole cards are displayed. The visual confirmation matches option (C), making it the correct choice. &
In the video, Hawk Tuah is shown holding a 2-pair hand in hand three. The cards she is holding are King of Hearts and Queen of Spades. This information is visible in the video at \textcolor{red}{around 20 to 25 seconds.} \\
\midrule
After the winner of the final hand stands up, where is the next person who stands up located in relation to the winner? &
To their left. &
I watched the winner of the final poker hand, Vegas Matt, standing at 07:18 while the AI narrator says that he won the hand. I then watched the video for the next person to stand up, which I saw at 07:20 as a woman stood and leaned forward over the table to gather poker chips. \textbf{I noted this woman's location in relation to Vegas Matt - she was directly to his left at the table.} &
In the video, after the final hand concludes and the winner, Vegas Matt, stands up, the next person to stand up is located to his left. This can be observed around the timestamp near the end of the video where the final hand is played. The seating arrangement shows that the person to the left of the winner is the next to stand, likely due to the elimination process in the tournament. The other options such as 'across the table', 'directly behind', 'to their right', or 'directly across the table' do not match the visual sequence of events captured in the video. &
\textcolor{red}{The video does not provide a clear view of the relative positions of the people standing up after the winner of the final hand.} However, based on the information provided in the video, it is possible to infer that the next person who stands up is located to the left of the winner. \\
\bottomrule
\end{tabular}}

\caption{\textbf{Qualitative Comparison of Model Reasoning:} We compare reasoning traces from InternVideo and Qwen2.5 for questions where both models got the correct multiple choice answer. InternVideo2.5 generally provides shorter reasoning traces with less evidence from the video (first to third examples) or simply restating the question and answer (second example). It also hallucinates (fourth example) or simply guesses when it fails to extract the evidence from the video (fifth example).}
\label{tab:reasoning_qualitative_comparison}
\end{table*}

\begin{table}[ht]
    \centering
    \footnotesize
\begin{tabular}{lllllllll}
\toprule
 & Qwen2.5-VL & VideoLLaMA3 & InternVideo2.5 & Claude 3.5 Sonnet v2 & OpenAI o1 (64f) & GPT-4.1 & \thead{Gemini 2.5 Pro\\(Thinking)} & Humans \\
\midrule
Misc Sports & 29.76 & 34.52 & 39.29 & 29.76 & 39.29 & 50.00 & 57.14 & 95.24 \\
Tennis & 27.87 & 36.07 & 27.87 & 32.79 & 44.26 & 36.07 & 55.74 & 95.08 \\
Travel & 30.77 & 23.08 & 25.96 & 27.88 & 35.58 & 45.19 & 65.38 & 91.35 \\
Motorsports & 41.67 & 35.42 & 37.50 & 43.75 & 56.25 & 68.75 & 72.92 & 95.83 \\
Tech/AI & 31.58 & 31.58 & 28.95 & 34.21 & 42.11 & 50.00 & 63.16 & 76.32 \\
Maths & 39.53 & 25.58 & 25.58 & 34.88 & 44.19 & 48.84 & 48.84 & 90.70 \\
Short Films & 48.14 & 48.55 & 47.11 & 44.01 & 54.13 & 68.60 & 76.65 & 97.31 \\
Basketball & 40.32 & 32.26 & 41.94 & 37.10 & 56.45 & 61.29 & 58.06 & 95.16 \\
Animals & 19.01 & 22.31 & 23.14 & 21.49 & 28.10 & 41.32 & 47.11 & 88.43 \\
Board Games & 34.04 & 31.91 & 29.79 & 26.24 & 39.72 & 44.68 & 65.96 & 95.74 \\
Physics & 16.33 & 30.61 & 32.65 & 24.49 & 34.69 & 38.78 & 61.22 & 81.63 \\
Chess & 21.82 & 32.73 & 27.27 & 33.64 & 35.45 & 47.27 & 56.36 & 92.73 \\
Cooking & 30.68 & 29.55 & 29.55 & 20.45 & 25.00 & 44.32 & 60.23 & 85.23 \\
How-To & 31.71 & 21.95 & 32.93 & 29.27 & 31.71 & 50.00 & 59.76 & 80.49 \\
\bottomrule
\end{tabular}
    \caption{MCQ scores by video category (see Fig.~\ref{fig:finegrained-acc}, middle)}
    \label{tab:mcq_scores_by_category}
\end{table}

\begin{table}[ht]
    \centering
    \footnotesize
\begin{tabular}{lllllllll}
\toprule
 & Qwen2.5-VL & VideoLLaMA3 & InternVideo2.5 & Claude 3.5 Sonnet v2 & OpenAI o1 (64f) & GPT-4.1 & \thead{Gemini 2.5 Pro\\(Thinking)}& Humans \\
\midrule
Temporal Reasoning & 32.88 & 35.16 & 33.03 & 32.27 & 40.33 & 51.29 & 61.19 & 93.15 \\
Counterfactual & 29.70 & 21.78 & 25.74 & 37.62 & 32.67 & 44.55 & 55.45 & 88.12 \\
Spatial Perception & 33.90 & 37.33 & 39.38 & 31.85 & 44.86 & 55.14 & 63.36 & 94.86 \\
Listening & 36.50 & 34.75 & 31.75 & 38.00 & 46.00 & 55.50 & 67.50 & 92.00 \\
State Changes & 34.43 & 34.43 & 36.07 & 32.79 & 42.62 & 57.38 & 59.02 & 88.52 \\
Event Occurence & 33.40 & 34.14 & 35.44 & 33.21 & 39.52 & 51.76 & 62.71 & 93.69 \\
Cause and Effect & 44.57 & 40.22 & 33.70 & 31.52 & 46.74 & 66.30 & 69.57 & 93.48 \\
Situational Awareness & 39.76 & 38.96 & 37.35 & 38.55 & 43.78 & 57.03 & 62.65 & 95.58 \\
Goal Reasoning & 45.00 & 40.00 & 42.50 & 32.50 & 50.00 & 62.50 & 77.50 & 92.50 \\
Numerical Reasoning & 26.82 & 27.20 & 24.14 & 32.57 & 38.31 & 41.38 & 59.39 & 87.36 \\
Counting & 24.07 & 25.19 & 28.89 & 27.78 & 30.37 & 40.00 & 51.67 & 86.11 \\
Reading & 39.23 & 34.92 & 33.56 & 35.37 & 47.17 & 56.46 & 70.98 & 92.52 \\
Object Recognition & 34.20 & 36.30 & 36.67 & 32.22 & 42.47 & 53.33 & 65.43 & 90.99 \\
\bottomrule
\end{tabular}
    \caption{MCQ scores by skill (see Fig.~\ref{fig:finegrained-acc}, left)}
    \label{tab:mcq_scores_by_skill}
\end{table}

\end{document}